%% file: main.tex
\documentclass{article}
\usepackage{iclr2027_conference_preprint,times}

\input{math_commands.tex}

\usepackage{hyperref}
\usepackage{url}
\usepackage{microtype}
\usepackage{graphicx}
\usepackage{cleveref}
\usepackage{booktabs}
\usepackage{multirow}
\usepackage{amsmath}
\usepackage[table]{xcolor}
\usepackage{subcaption}
\usepackage{placeins}
\usepackage{comment}
\usepackage{wrapfig}
\usepackage{pgf}
\usepackage{needspace}

\DeclareGraphicsExtensions{.png,.jpg,.pdf,.ai,.psd}
\definecolor{best}{RGB}{255,140,140}   % red
\definecolor{second}{RGB}{255,190,120} % orange
\definecolor{third}{RGB}{255,225,140}  % yellow
\renewcommand{\arraystretch}{1.10}
\usepackage{siunitx}

\usepackage{acronym}
\acrodef{3DGS}{3D Gaussian Splatting}
\acrodef{SLAT}{structured latent}
\acrodef{VAE}{variational autoencoder}
\acrodef{FID}{Fréchet Inception Distance}
\acrodef{KID}{Kernel Inception Distance}

\def\clap#1{\hbox to 0pt{\hss#1\hss}}
\def\initials#1{%
  \protect\clap{\smash{\raisebox{1.4ex}{%
    \tiny\textsf{\textit{~~#1}}}}}}

\makeatletter
\newcommand{\NOTE}[3]{%
  \protect\@ifundefined{hidecomments}{%
    \strut{\color{#2}\initials{#1}%
    \protect{{\small$\lfloor$}#3{\small]}}}%
  }{}%
}
\definecolor{nicegreen}{rgb}{0,0.65,0.2}
\makeatother

\definecolor{ElieColor}{RGB}{170,46,200}
\definecolor{NoeColor}{RGB}{224,128,65}
\newcommand{\MF}[1]{\NOTE{MF}{blue}{#1}}
\newcommand{\EM}[1]{\NOTE{EM}{ElieColor}{#1}}
\newcommand{\NL}[1]{\NOTE{NL}{NoeColor}{#1}}

\newcommand{\elieUpdate}[1]{{\textcolor{ElieColor}{#1}}}

\renewcommand{\MF}[1]{}
\renewcommand{\EM}[1]{}
\renewcommand{\NL}[1]{}

\renewcommand{\elieUpdate}[1]{{#1}}

\newcommand\pgl{\mathcal{P}_{\mathcal{G}_L}}
\newcommand\z{\boldsymbol{z}}
\newcommand\gaussianmask{$\mathcal{M}$ }

\newcommand\wing{\includegraphics[height=0.9em]{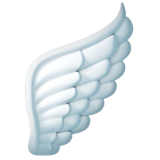}}
\newcommand\wingfoot{\includegraphics[height=1.2em]{figures/wing.png}}

\title{WINGS\wing: Reference-Free Gaussian Splatting \\ Inpainting with 3D-Native Generative Priors}

\iclrpreprint

\author{Noé Lallouet \\
Adobe \\
MILES Team, LAMSADE \\
Paris Dauphine -- PSL University \\
Paris, France \\
\texttt{noe.lallouet@dauphine.eu} \\
\And
Michael Fischer \\
Adobe\\
London, UK \\
\texttt{mifischer@adobe.com} \\
\And
Elie Michel \\
Adobe\\
Paris, France \\
\texttt{emichel@adobe.com} \\
}

\begin{document}
\maketitle

{
    \renewcommand{\thefootnote}{\raisebox{-0.4em}{\wingfoot}}
    \footnotetext{\textbf{W}e \textbf{IN}paint \textbf{G}aussian \textbf{S}cenes.}
}

\begin{abstract}
Inpainting 3D Gaussian Splatting scenes, a key challenge in 3D editing, requires generating plausible content within a masked region of 3D space.
Prior approaches rely on 2D diffusion models to produce one or several inpainted reference views, making them susceptible to challenges associated with multi-view inconsistency and lengthy optimization times.
Departing from these approaches, we introduce a reference-free Gaussian splatting inpainting method operating natively in 3D. 
Our method leverages the embedding space of a large, pre-trained 3D prior, combined with a structure completion network to feed a generative prior which reconstructs the missing region's geometry and appearance.
Performing content generation entirely in 3D, it avoids the need to reconcile inconsistencies of multiple inpainted reference images, and is faster than related 2D-based methods.
We demonstrate the effectiveness of our method qualitatively and quantitatively, through extensive experiments and a user study. 
To the best of our knowledge, this work is the first Gaussian splatting inpainting method to operate in the learned representation space of a 3D-native generative prior without relying on inpainted reference views.
\end{abstract}

%%%%%%%%%%%%%%%%%%%%%%%%%%%%%%%%%%%%%%%%%%%%%%%
\section{Introduction}

\ac{3DGS}~\citep{kerbl_3d_2023a} has become the de facto standard for scene reconstruction and novel view synthesis due to its high fidelity, explicit representation and efficient rendering.
As a consequence, many recent works focus on editing \ac{3DGS} assets and scenes, often with the goal of changing or removing objects from the scene. 
Object removal, specifically, leads to an interesting problem: parts of the scene that were previously hidden (e.g., the tabletop on which a vase had rested before removal) become disoccluded and are now visible, but were neither observed nor reconstructed during the initial \ac{3DGS} training, leading to empty holes and artifacts.
Such areas now need to be inpainted: an algorithm needs to synthesize a plausible 3D continuation of surrounding surfaces and appearance that remains coherent under novel views, a problem that becomes especially difficult when the original 2D observations are no longer available. 

%\EM{Highlight from the beginning that we operate on gsplat that do not necessarily come from scanning pipelines/reference images.} \MF{Added to prev. paragraph, but given that our LORA still req rendered views I wouldnt put too much emphasis on it tbh}

Earlier existing \ac{3DGS} inpainting methods use 2D image diffusion models to provide one or several 2D-inpainted reference views of the scene, which are then optimized back into the 3D representation through differentiable rendering. 
They therefore solve a 3D-completion problem indirectly, by first hallucinating 2D projections and subsequently optimizing a 3D representation to explain them. 
Since the generative prior operates in 2D, the inpainted content often is not multiview-consistent, and it requires special care to consolidate these disagreements into 3D, often leading to blurry artifacts in and around the inpainted regions.
Additionally, generating multiple inpainted reference views and re-optimizing the scene introduces significant computational overhead, leading to long inpainting times~\citep{wu_amodal3r_2025a} and motivating the reduction of this complexity by operating directly in 3D.

Recent large-scale 3D generative models provide a compelling alternative: rather than generating 2D content that is subsequently lifted into 3D, their learned priors enable reasoning about the missing content directly in 3D.
Motivated by this observation, we introduce a reference-free approach to 3DGS inpainting that operates natively in three-dimensional space. 
%To this end, we perform scene-specific inpainting by leveraging the representational capabilities of a large-scale pretrained 3D generative prior. \MF{the before sentence feels disconnected}. 
Specifically, we start by embedding the 3DGS scene into the latent space of the 3D prior. 
Expressing the observed 3D context in a latent space, we predict the missing geometric structure of the target region, providing spatial support for the following latent masked region feature generation.
Having predicted the target region's structure, we subsequently generate the associated appearance features by conditionally sampling from the generative prior, via a combination of a progressive context-conditioning algorithm and a novel self-attention masking mechanism. 
Finally, we transform the generated latent features to Gaussians via a generative decoder adapted to the observed 3DGS scene in order to ensure perceptual consistency between existing and generated content, before merging generated and existing context Gaussians.

%\MF{v1, too much architectural details, too little high-level motivation:}
%Given a masked region where an object has been removed and a 3DGS scene, we embed them into the \ac{SLAT} embedding of a large, powerful pre-trained 3D-native genererative model.
%We mask the structured latents and run a sparse structure completion network to recover the inpainted region's coarse structure by sampling plausible Gaussians.
%We then run a conditional flow model with attention masking in order to inpaint the missing latents in 3D embedding space, before decoding the inpainted embedding back into explicit 3DGS primitives that seamlessly merge back into the scene, without requiring specific optimization. 
Importantly, our approach does not require inpainted reference views from a 2D model: the missing content is generated directly in 3D space, which avoids blur and artifacts from multi-view inconsistencies of several hallucinated 2D completions. 
Further, generating content in 3D allows us to skip expensive post-inpainting 3DGS optimization, which notably reduces inference time, a key requirement in interactive applications. 
The generated Gaussians are, by design, multiview-consistent, and conditioning the generation on the observed context and adapting the decoder to scene-specific characteristics allows the generated Gaussians to seamlessly integrate with the existing scene.

The contributions of this work are the following: \textbf{(i)} we introduce \textbf{\textsc{Wings}}: a novel, reference-free framework for inpainting 3DGS with a 3D-native generative prior that does not rely on any 2D-inpainted reference views.
%Contrary to existing approaches, we do not rely on inpainted reference views and perform structure completion and appearance generation in a structured 3D latent space instead. 
\textbf{(ii)} We propose a novel structure-completion network and an attention-masking strategy for structured latent inpainting that 
%combined with an existing context conditioning mechanism, 
enables generation of context-consistent target-region features at inference time without the need to retrain the generative prior.
\textbf{(iii)} We introduce a scene-conditioned decoder adaptation procedure that bridges the representation gap between the learned prior and the observed scene's arbitrary appearance distribution. 
%, thus improving perceptual consistency between existing and generated content.

% \begin{itemize}
%     \item We introduce a novel reference-free framework for inpainting Gaussian splatting scenes with a 3D-native generative prior. Contrary to existing approaches, we do not rely on inpainted reference views, instead performing structure completion and appearance generation in a structured 3D latent space.
%     \item We propose a self-attention masking strategy for structured latent inpainting that, combined with an existing context conditioning mechanism, enables generation of context-consistent target region features at inference time without the need to retrain the generative prior.
%     \item We introduce a scene-conditioned decoder adaptation procedure, in the form of a lightweight adjustment of the generative decoder with respect to the input 3DGS scene, bridging the representation gap between the learned prior and the observed scene's distribution, thus improving perceptual consistency between existing and generated content.
% \end{itemize}

%\MF{contributions part feels a bit repetitive now}

%%%%%%%%%%%%%%%%%%%%%%%%%%%%%%%%%%%%%%%%%%%%%%%
\section{Related Work}

% \subsection{3D Gaussian Splatting Inpainting}
% \label{subsec:rw_3dgs_inpainting}

% Just listing methods with a 1-sentence description for now, will compact later

\textbf{3D Gaussian Splatting Inpainting.} The first methods for 3D radiance field inpainting, SPIn-NeRF~\citep{mirzaei_spinnerf_2023a} and \citet{weder_removing_2023}, address object removal in NeRFs by obtaining multiview segmentation masks and independently inpainting the masked RGB input views with a learned 2D inpainter. 
To mitigate inconsistencies across the inpainted views, SPIn-NeRF optimizes the NeRF via a perceptual loss rather than pixel-wise supervision, and further inpaints depth-maps from the original NeRF to use as geometric priors during optimization. %, yielding an inpainted NeRF that is consistent across novel views.
Adapting this idea to Gaussian splatting, a first category of 3DGS inpainting methods leverages multiple views inpainted through a 2D diffusion model followed by optimization from points initialized around the target region. 
GaussianEditor \citep{chen_gaussianeditor_2024} transforms an inpainted render to a mesh, which is then re-converted to Gaussians and re-optimized with supervision from the inpainted view. 
Gaussian Grouping~\citep{ye_gaussian_2024a} inpaints in 2D with LaMa~\citep{suvorov_resolutionrobust_2022}, then lifts inpainted pixels to 3D and proceeds to 3DGS optimization supervised by the inpainted views.

More recently, 3DGS inpainting approaches have largely relied on joint appearance and depth inpainting. 
GScream~\citep{wang_learning_2024} uses guidance from estimated monocular depth to initialize Gaussians from a single 2D inpainted view, while GPGS~\citep{lee_gpgs_2026} leverages a point cloud masked autoencoder to perform geometry completion. 
AuraFusion360~\citep{wu_aurafusion360_2025a} uses depth diffusion to initialize Gaussians, and Inpaint360GS~\citep{wang_inpaint360gs_2026} performs depth-guided inpainting, combined with cross-frame conditioning in order to improve multi-view consistency. 
3DGIC~\citep{huang_3d_2025a} uses a similar depth-guided approach to identify background pixels.
Similarly, InFusion~\citep{liu_infusion_2024a} combines inpainted reference views with a depth completion model, and~\citet{zhou_highfidelity_2025a} proposes an uncertainty-guided approach to Gaussian inpainting.

Recent approaches increasingly investigate inpainting in 3D space directly, with or without inpainted reference views. InstaInpaint~\citep{you_instainpaint_2025a} leverages a large reconstruction model~\citep{zhang_gslrm_2025} to inpaint a 3DGS scene conditioned on a unique inpainted reference view. 3D-GIMP~\citep{tian_3dgimp_2026} similarly inpaints a single keyframe in 2D and then runs a 3D-aware patch-match algorithm, while GS-RoadPatching~\citep{chen_gsroadpatching_2025} searches for correspondences in 3D with applications to completion of road scans.  

Despite increasingly incorporating 3D-aware geometric priors, many of the aforementioned approaches still derive the missing region's appearance from one or more 2D inpainted reference views and subsequently reconcile or refine this information in the 3D representation via depth-cues, cross-view-attention or other auxiliaries. 
Our method, in contrast, operates directly in 3D space.

\textbf{Generative priors.} Recently, generative models have emerged as remarkable means of creating 3D assets with text or image conditioning \citep{xiang_structured_2025a, xiang2025trellis2, zhao_hunyuan3d_2026, yan_generative_2026c}. While largely operating inside learned latent spaces, some generative 3D models interface with 3D Gaussians, through variational autoencoders~\citep{xiang_structured_2025a} or direct generation~\citep{yan_generative_2026c}.

VoxHammer~\citep{li_voxhammer_2026} performs inpainting and editing on 3D structured latents by sampling from a general 3D generative model~\citep{xiang_structured_2025a}, relying on a single inpainted image injected as conditioning in the prior. 
%Albeit related to our method, VoxHammer is limited to mesh- and structured-latent editing and does not apply to 3DGS, while requiring an inpainted reference view.
Albeit related to our method, VoxHammer requires an inpainted reference view, is focused on mesh- and structured-latent-editing, and restricts 3DGS applications to asset editing.
Similarly, InpaintSLat~\citep{chung_inpaintslat_2026} proposes a structured latent inpainting framework based on the TRELLIS model~\citep{xiang_structured_2025a} to edit 3D assets. 
Importantly, both frameworks focus on mesh editing and object-level completion, rather than full-scene inpainting. 

Overall, existing \ac{3DGS} inpainting methods overwhelmingly rely on 2D diffusion priors, while 3D-native editing methods largely operate within the learned distribution of their native representations, whose appearance domain often differs substantially from that of photorealistic, reconstructed scenes. 
Bridging the two, we show how a powerful pre-trained 3D generative model can be transferred to arbitrary Gaussian splatting scenes by conditionally generating a latent representation and adapting the generative decoder to the scene-specific, photorealistic Gaussian distribution.

\textbf{Training-Free Conditioning of Generative Priors.}
A complementary line of work repurposes pre-trained generative models for conditional generation without task-specific re-training. RePaint~\citep{lugmayr_repaint_2022a} performs image inpainting with an unconditional diffusion model by repeatedly injecting observations of the known region throughout the reverse process, and harmonizes generated with observed content via re-sampling. 
More generally, diffusion posterior sampling (DPS, ~\cite{chung2023diffusion} incorporates measurement constraints into generation through likelihood-guided updates, while ReSample~\citep{song_solving_2023} enforces hard data consistency when solving inverse problems with latent diffusion models. Orthogonally, inference-time attention control methods such as MasaCtrl~\citep{cao_masactrl_2023a} manipulate self-attention to control information flow during generation without modifying the underlying model. 
~\citet{sartor_overpainting_2026} further show that joint attention with attention dropout can be leveraged to control interactions between input modalities. 

In contrast to the previous methods, our approach builds on these principles in a structured \emph{three-dimensional} latent space, conditioning a frozen 3D generative prior on observed scene latents while restricting information flow between observed and inpainted regions.

% RePaint / Resample 
% MasaCTRL 
% DPS 
% GPGS ok
% VoxHammer ok
% SPIN-NeRF ok 
% GScream ok
% AuraFusion360 ok
% GauGrouping ok
% Inpaint360GS ok

%%%%%%%%%%%%%%%%%%%%%%%%%%%%%%%%%%%%%%%%%%%%%%%
\section{Method}

% General challenges:
% - multiple plausible inpainting
% - aliasing of the GSplat representation
% => P(G_m|G_c) is highly multimodal, hence hard to sample
% We reuse pre-trained prior. A 2D prior would enable sampling P(Full Image|Render(G_c)) but on multiple views this conflicts with the high multimodality of the problem. We instead rely on a 3D-native (unconditional) prior P(G).

% N: Changed the Gaussian mask to $\mathcal{M}, we are ready to switch the attention mask back to \gaussianmask. 
% Removed $G_m$ in the first sentence because we say 2 lines lower that we want to predict G_m, so the masked removed Gaussians and the masked region generated Gaussians had the same symbol. Since we delete them, they don't need a symbol
% E: Ok, FYI I propagated the change of M->\gaussianmask and U->M to the rest of the text
We consider a 3DGS scene $G$ and a user-prescribed masked region \gaussianmask. After removing the Gaussians lying in \gaussianmask, usually corresponding to an unwanted object, our goal consists in generating Gaussian primitives $G_m$ inside the masked region while remaining consistent with the observed remaining context $G_c$. Adopting the probabilistic view of \cite{kheradmand_3d_2024}, we thus aim at sampling masked region Gaussians from the conditional distribution $G_m \sim \mathcal{P}(G_m | G_c)$.

This conditional distribution is highly multi-modal at two levels. At a semantic level, the generation of plausible content admits infinitely many valid solutions, and at the primitive level, multiple combinations of Gaussians can result in perceptually identical results. Because of the former, relying on 2D inpainting priors~\citep{ye_gaussian_2024a, liu_infusion_2024a} can lead to multi-view inconsistencies; while the latter presents a significant obstacle to generating content directly in the space of 3D Gaussians. Motivated by these observations, our method operates in the structured latent space of a pre-trained 3D-native generative prior, namely TRELLIS~\citep{xiang_structured_2025a}.

\begin{figure}
    \centering
    \includegraphics{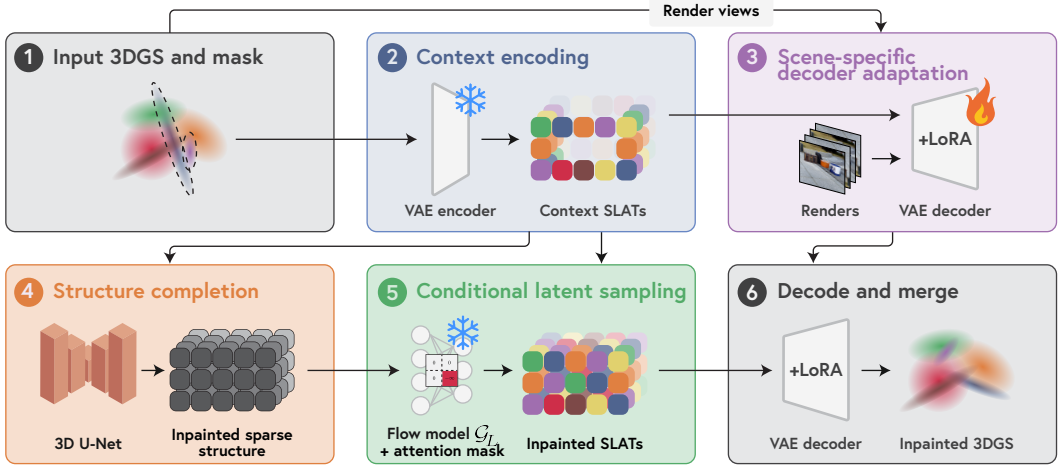}
    \vspace{-5mm}
    \caption{Overview of \textsc{Wings}. Given the input 3DGS encoded as \ac{SLAT} and a mask $\mathcal M$, our structure completion network predicts masked-region occupancy. 
    The \ac{SLAT}s are inpainted directly in 3D, conditioned on the context latents and decoded to Gaussian splats.
    }
    \label{fig:diagram}
\end{figure}

After recalling the general architecture of the 3D-native generative prior used here (\Cref{subsec:background}), we describe our protocol for sampling $G_m \sim \mathcal{P}(G_m | G_c)$. First, we predict the sparse voxel structure of the missing region (\Cref{subsec:encoding_and_structure_completion}), before generating latent features for these voxels through a combination of conditional sampling and a self-attention masking mechanism (\Cref{subsec:conditional-sampling}). Finally, we adapt the latent decoder to the specific appearance of the scene (\Cref{subsec:scene_conditioned_decoding}), and merge the decoded Gaussians with the original context. An overview of our method is displayed in \Cref{fig:diagram}.

\begin{comment}
Given a 3DGS scene consisting of Gaussians $G$ and a 3D mask \gaussianmask, our objective is to generate plausible Gaussians inside the masked region while remaining consistent with the surrounding context. 
Our method operates through the latent representation of a pre-trained 3D generative model.
We first encode the neighborhood of the target region through a frozen VAE encoder, obtaining a latent context representation.
We then predict the sparse voxel structure of the missing region (\Cref{subsec:encoding_and_structure_completion} and generate features for the predicted voxels through a combination of conditional sampling and a self-attention masking mechanism (\Cref{subsec:conditional-sampling}). 
Finally, the VAE decoder is adapted to the specific appearance of the scene (\Cref{subsec:scene_conditioned_decoding}), and the generated latents are decoded to Gaussians and merged with the original scene. 
\Cref{fig:diagram} shows an overview of our method.
\end{comment}

%----------------------------------------------
\subsection{Pretrained 3D-native prior}
\label{subsec:background}

The generative prior we use operates in a learned latent space, where 3D content is modeled by a collection of $L$ $d$-dimensional latent tokens $\z \in \mathbb{R}^{d \times L}$ obtained by encoding the Gaussian splats. 
These latents reside on a % $64^3$ occupancy 
voxel grid, and are henceforth denoted as \emph{structured} latents (\ac{SLAT}s), enabling us to distinguish the latents $\z_c$ obtained from the observed context from the latents $\z_m$ that we generate within the masked region $\mathcal{M}$: we denote $\z = \z_m \cup \z_c$. Our goal is thus to sample from the conditional distribution $\mathcal{P}_{\mathcal{G}_L}(\z_m | \z_c)$, which is induced by a latent generator $\mathcal{G}_L$.

In practice, we use the two-stage TRELLIS architecture \citep{xiang_structured_2025a}: a first model $\mathcal{G}_S$ predicts the latent occupancy structure $O$ (a $64^3$ binary voxel grid) and a second model $\mathcal{G}_L$ assigns latent features $z_i$ to active voxels through rectified flow integration. We use TRELLIS.1 as it provides a pre-trained variational autoencoder interfacing Gaussians with \ac{SLAT}s. A notable challenge that we address in this paper is inpainting Gaussians from photorealistic, multi-object scenes, which conflicts with the non-photorealistic appearance of TRELLIS-generated assets stemming from the model's object-centric training procedure. We tackle this distribution mismatch through inpaint-specific structure completion (\Cref{subsec:encoding_and_structure_completion}),  conditional latent sampling (\Cref{subsec:conditional-sampling}) and scene-conditioned decoder adaptation (\Cref{subsec:scene_conditioned_decoding}).

\subsection{Encoding and structure completion}
\label{subsec:encoding_and_structure_completion}
    
\paragraph{Context encoding.} Although we seek to perform inpainting on potentially large scenes, the TRELLIS prior operates at a specific scale determined by the fixed resolution of its \ac{SLAT} structure. 
We therefore extract a local context $G_{c^*} \subset G_c$ around the target region \gaussianmask by discretizing and finding the smallest bounding box such that the total ratio of voxels in $G_{c^*}$ to voxels in the masked region is $6$. 
We find this empirically determined ratio to be a good trade-off between enough context information and sufficiently dense resolution for the reconstructed Gaussians $G_m$.
Our early experiments revealed the TRELLIS latent flow model $\mathcal G_L$ to be biased towards generating axis-aligned structures, so we automatically align the bounding box with the visual features of the context $G_{c^*}$ using the structure tensor \citep{bigun_multidimensional_1991} (see Appendix~\ref{appendix:subsec:autoalign} for details).
Following the TRELLIS protocol, we finally encode this context into context latents $\z_c$ via the pretrained VAE encoder $\mathcal{E}$. 

\paragraph{Structure completion.}
The $64^3$ binary occupancy grid $O$ is known for context but initially empty for the voxels corresponding to the masked region \gaussianmask\!. 
Accurate completion of this structure is crucial, since the subsequent latent generator $\mathcal G_L$ can only produce content for voxels marked as active in $O$. \citet{xiang_structured_2025a} use a structure flow model $\mathcal G_S$ for this task and investigate the application of the RePaint algorithm~\citep{lugmayr_repaint_2022a} for structural completion. 
However, we find that $\mathcal G_S$'s learned objective of generating new, complex content competes with the inpainting task, which aims at producing a visually plausible continuation of the surrounding context rather than complex new structures. While satisfactory for image-conditioned generative inpainting tasks~\citep{xiang_structured_2025a, li_voxhammer_2026}, $\mathcal{G}_S$ exhibits degraded generation quality in the structure completion regime in which we operate (see \Cref{fig:repaint-vs-unet} for an analysis).

We therefore opt to train a new structure completion model that directly predicts $O$ based on the context structure $O_c$ and the target region mask \gaussianmask\!.
We choose a 3D residual U-Net architecture \citep{cicek_3d_2016a, he_deep_2016} with bottleneck attention, which allows us to benefit from the translation-equivariance and multi-scale receptive field of convolutional networks, while preserving fine-grained spatial structure through skip-connections. We train our model on a curated corpus of 900 diverse voxelized scene-level 3D Gaussian splatting scenes with randomly sampled masked regions. No scene from any of our benchmark evaluation datasets was used in training the structure prediction network. In order to guarantee clean structure completion example pairs, we curate the dataset by rejecting samples where the target structure deviates substantially from a quadric extrapolation of the context structure or contains high volumetric variations. The dataset curation process is further detailed in Appendix~\ref{appendix:subcsec:structure-dataset} and results in more than 71,000 training samples. We train our structure completion network for 300 epochs using a linear combination of weighted binary cross-entropy and Dice loss:
% $
%     \mathcal{L}_\text{structure} = \alpha \cdot \mathcal{L}_\text{W-BCE} + (1 - \alpha) \mathcal{L}_\text{DICE} \,
% $ (with $\alpha = 0.3$).
$
    \mathcal{L}_\text{structure} = 0.3 \, \mathcal{L}_\text{W-BCE} + 0.7\, \mathcal{L}_\text{DICE} \,.
$

\subsection{Conditional structured latent sampling} \label{subsec:conditional-sampling}

The predicted structure $O_m$ specifies the spatial structure of target region latents $z_m$, but not the value of their latent features, which we must sample from the conditional distribution $\mathcal P_{\mathcal G_L}(\z_m | \z_c)$. 
However, TRELLIS' flow model $\mathcal G_L$ only allows us to sample from the \emph{unconditional} prior $\mathcal P_{\mathcal G_L}(\z)$ (notwithstanding image/text conditioning injected through cross-attention, unused here). In the following section, we show that this general prior can be used to approximate conditional sampling in the case of inpainting by combining two schemes: context conditioning and self-attention masking.

\begin{comment}
With the SLAT structure now sufficiently refined, we turn to the problem of assigning visually plausible features to the generated SLATs. 
As stated earlier, generating Gaussians in the masked region requires sampling from the conditional probability distribution $\mathcal{P}(G_m | G_c)$: the inpainted Gaussians must be consistent with the context surrounding them. \MF{inconsistent: previously we were talking about things in latent space, yet here we mention Gaussians in 3D space} \NL{Agree, will change}
To this end, we leverage two training-free mechanisms to approximate a conditional sampler for inpainting from the general prior. 
We emphasize that the latent flow model, in our framework, requires no image or text conditioning, such as a reference image inpainted with a 2D diffusion model; generation runs unconditionally, and sampling from the conditional distribution is made possible by the two following levers:
\end{comment}

\paragraph{Progressive context conditioning.}

%\NL{\EM{Make it more explicit that our contribution doesn't lie here} Maybe use stronger vocabulary than "inspired". Elie suggests making this section shorter, we must balance the lack of novelty here with the fact that it's an essential pipeline component}

%Inspired by TRELLIS~\cite{xiang_structured_2025a}, we first adopt the RePaint conditioning strategy~\cite{lugmayr_repaint_2022a} to sample masked region latents. Importantly, this strategy does not require a conditional flow model: conditional generation is achieved by altering the sampling trajectory. 
The conditional sampling scheme that we adopt is based on the RePaint strategy~\citep{lugmayr_repaint_2022a}, which repurposes an unconditional flow model by altering the sampling trajectory.
Let $\z^t = \z_m^t \cup \z_c^t$ be the structured latents at flow time $t$. 
At each integration step, the flow model $\mathcal{G}_L$ 
predicts a rectified flow velocity $v(\z^{t}, t)$  and advances through the flow trajectory by generating $\z^{t+dt} = \z^{t} + v(\z^t, t) dt$. Given that the true context latents $\z^0_c$ are known, RePaint then replaces the context $\z^{t+dt}_c$ with $\hat{\z}^{t+dt}_c$, the known context latents noised at that time:
\begin{equation} \label{eq:repaint}
    \z^{t+dt} = \hat{\z}^{t+dt}_c \cup \z^{t+dt}_m \quad\text{ where }\quad
    \hat{\z}^{t}_c = (1 - t)\z^0_c + t \epsilon, \quad \epsilon \sim N(0,1).
\end{equation}
In order to harmonize the joint generation of context and masked latents, this consistency replacement is supplemented by stochastic renoising jumps that provide additional opportunities for the masked region to be denoised towards context-consistent latents. While this conditional sampling strategy was originally suggested by TRELLIS~\citep{xiang_structured_2025a}, its efficacy is highly tied to the use of their original structure prediction model $\mathcal G_S$, as its generated sparse structure is generally in-distribution for  $\mathcal G_L$. 
While switching to our U-Net predictor improves structure completion, it also creates a potential mismatch between the context and masked region distributions.
Using a high number of flow integration timesteps for SLAT generation alleviates this issue, but leads to a significant slowdown. 
We instead introduce a complementary conditioning mechanism that alters the flow transformer's self-attention.

\paragraph{Self-attention masking.}

\begin{figure}
    \centering
    \includegraphics{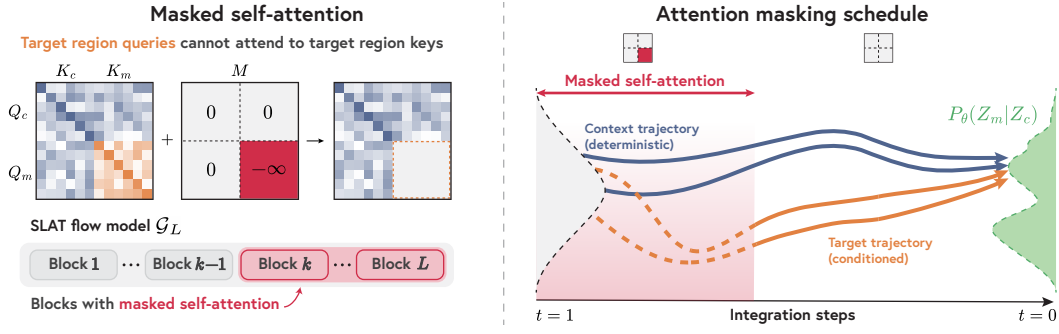}
    \vspace{-6mm}
    \caption{Attention masking. The attention mask acts on the query-key dot product $Q K^\top$, effectively zeroing out target-to-target self-attention interactions. We note that, since the mask is not applied to the value tensor $V$, the output of self-attention is not null for target latents.}
    \label{fig:attention-masking}
\end{figure}

The TRELLIS latent flow model $\mathcal{G}_L$ uses transformer blocks that compute multi-head self-attention over structured latents. Applying the RePaint strategy here initially leaves the target region latents $\z_m$ primarily determined by the flow model's unconditional trajectory. 
Since they continuously self-attend throughout the transformer and due to potential sparse structure distributional mismatch, they risk to form a self-consistent region that is too robust for the context to impose its condition.We therefore propose to alter the information flow during integration in a way that explicitly promotes information sharing between context and target latents, while simultaneously hindering their self-attention. 
Inspired by the mutual self-attention mechanism from~\cite{cao_masactrl_2023a}, we use an additive attention mask $M$ to prevent the target region's queries from attending to any of the target region's keys:
\begin{equation}
    A_{\text{masked}} = \text{softmax} \left( \frac{QK^\top}{\sqrt{d_k}} + M \right), \quad M_{ij} = \begin{cases}
        - \infty & \text{if } z_i \text{ and } z_j \text{ are in } \z_m \\ 0 & \text{otherwise.}
    \end{cases}
\end{equation}
This is complementary to RePaint, as the latter enforces conditioning through backwards jumps along the sampling trajectory, stochastic re-noising and context latent replacement at each integration time step, while our attention masking mechanism directly alters information flow during the evaluation of the velocity $v(\z^{t}, t)$. 
In order to preserve self-consistency within the masked region, we apply this masked self-attention only for later transformer blocks (see Appendix~\ref{appendix:subsec:implem}) and only for integration steps between $t=1.0$ and $t=0.8$ (see \Cref{fig:attention-masking}). In summary, combining RePaint-like jumps with our masked attention strategy allows us to conditionally sample the unconditional flow model $\mathcal{G}_L$ and generate latents that are conditioned on, and therefore consistent with, the context region.

\subsection{Scene-conditioned decoding}
\label{subsec:scene_conditioned_decoding}

After having obtained structured target latents $z_m$ via the aforementioned conditional latent sampling stage (\Cref{subsec:conditional-sampling}), we decode $\z = \z_m \cup \z_c$ into Gaussians using the pre-trained VAE decoder $\mathcal{D}_\theta$ with parameters $\theta$. However, since $\mathcal{D}_\theta$ was trained on the limited, object-centric dataset of TRELLIS, its induced pushforward measure $(\mathcal{D_\theta)}_\#(\pgl (\z_m | \z_c))$ is not guaranteed to match our target $\mathcal P(G_m | G_c)$. As such, even semantically correct latents can be decoded into perceptually dissimilar Gaussians, resulting in visually unpleasant boundary discontinuities between context $G_c$ and generated Gaussians $\mathcal{D}_\theta(\z_m)$.

We address this distribution mismatch by adapting the decoder to the observed scene's appearance. Alongside the latent generation process, we augment $\mathcal{D}_\theta$ with a rank-16 LoRA adapter \citep{hu_lora_2021a} for $\mathcal{D}_\theta$. We unfreeze the final layer of $\mathcal{D}_\theta$, responsible for generating the Gaussian opacity, scale, and rotation values. The 2.70 M parameters $\phi$ of this lightweight adapter are optimized through minimization of a photometric loss on rendered views:
\begin{equation}
    \phi^* = \arg \min_\phi \underset{v \sim \mathcal{V}}{\mathbb{E}} \Big[ \mathcal{L}_{\text{data}} \big( \mathcal{R}(\mathcal{D}_{\theta, \phi}(\z_c), v); \mathcal{R}(G_{c^*}, v)   \big)\Big]
\end{equation}
where $\mathcal{R}$ is a differentiable rasterizer and $\mathcal{V}$ is a collection of 36 anchor viewpoints. %\elieUpdate{Although this stage of our pipeline relies on rendered views, this does not impact the cross-view consistency of our results since what we optimize (the decoding $\mathcal D$) does not depend on the viewpoint. 
Importantly, these renders only provide supervision for adapting the view-independent decoder to the scene; they are never 2D-inpainted or serve as inputs or conditioning for the completion process. 
The same adapted decoder and latent representation are shared across all viewpoints. 
We also do not require specific reference views as we sample synthetic viewpoints around the cropped context $G_{c^*}$.

\section{Results}
\label{sec:results}

We evaluate our method across three 360° radiance-field inpainting benchmarks using both quantitative image-quality metrics and a user study, as well as qualitative analysis.

\begin{comment}
We evaluate our method against three recent 3DGS inpainting approaches: Gaussian Grouping \citep{ye_gaussian_2024}, InFusion \citep{liu_infusion_2024a}, and Inpaint360GS \citep{wang_inpaint360gs_2026}. 
We assess performance using quantitative image-quality metrics, a large-scale \EM{"large scale" is a bit too subjective -> tone down} user study, and qualitative comparisons across three 360° scene-inpainting benchmarks. \EM{To me, "360° scene" means a panoramic photography, not 3DGS, is this the term other papers use?}
\end{comment}

\textbf{Baselines and evaluation protocol.}
We compare against Gaussian Grouping \citep{ye_gaussian_2024a}, InFusion \citep{liu_infusion_2024a}, and Inpaint360GS \citep{wang_inpaint360gs_2026}, the latter of which reports strong performance on its proposed Inpaint360 benchmark. 
For all methods, we use the same input 3DGS scene and pre-defined 3D mask. 
This differs from evaluation protocols in which scene reconstruction and mask segmentation are performed independently for each method \citep{wu_aurafusion360_2025a,wang_inpaint360gs_2026}, as we intend to isolate differences due to the inpainting method itself. As such, our results also apply to 3DGS scenes that do not result from an optimization-based reconstruction. 
All experiments run on a single NVIDIA H100 GPU; see \Cref{appendix:subsec:implem} for implementation details.

\textbf{Datasets.}
We evaluate on three unbounded 360° radiance-field inpainting datasets: (1) COR-NeRF \citep{lee_consistent_2024}, containing 11 diverse indoor and outdoor scenes with objects at different scales; (2) Inpaint360 \citep{wang_inpaint360gs_2026}, containing 11 single- and multi-object scenes; and (3) 360-USID \citep{wu_aurafusion360_2025a}, containing seven scenes. In total, our evaluation comprises 29 scenes.

\subsection{Quantitative and qualitative results}

\textbf{Automated metrics.} We evaluate \textsc{Wings} on a comprehensive collection of quantitative metrics. We employ distributional perceptual metrics as our main quantitative evaluators: for each method, we report the \ac{FID} \citep{heusel_gans_2017} and \ac{KID} \citep{binkowski_demystifying_2018}. As \ac{FID} resizes images to $299\!\times\!299$ to accommodate the Inceptionv3 backbone's input size, it is poorly suited for evaluating fine detail in the inpainted region \citep{xie_smartbrush_2023, pavasovic_wait_2026}. To address this, we additionally report Local-FID, obtained by cropping a $299\!\times\!299$ region centered on the target region, thus allowing FID to operate at full resolution. We also report LPIPS.
%Our choice of metrics, particularly the omission of PSNR and SSIM despite being strongly competitive with existing methods here, is detailed in appendix~\ref{appendix:subsec:additional-results}. 
As detailed in Appendix~\ref{appendix:subsec_additional-results}, we set aside PSNR and SSIM since such reference-based metrics do not account for the large number of valid inpainting solutions in the target region.

\begin{wraptable}[8]{r}
{0.67\linewidth}
\vspace{-\intextsep}
    \centering
    \small
    \caption{User study. Each cell contains the tie-adjusted win-rate (higher is better) in \%, row-vs-column. See App.~\ref{appendix:user-study} for details.}
    \label{tab:user-study-global}
    \vspace{-2mm}

    \setlength{\tabcolsep}{3pt}
    \resizebox{\linewidth}{!}{%
\begin{tabular}{lccccc}
\toprule
Method & Ours & Inpaint360GS & Gaussian Grouping & InFusion & Expected rank $\downarrow$ \\
\midrule
Ours & \cellcolor[HTML]{DDDDDD}-- & \cellcolor[rgb]{0.594,0.765,0.634}75.11 & \cellcolor[rgb]{0.498,0.730,0.552}84.19 & \cellcolor[rgb]{0.404,0.697,0.473}93.01 & \textbf{1.48 $\pm$ 0.07} \\
Inpaint360GS & \cellcolor[rgb]{0.830,0.519,0.574}24.89 & \cellcolor[HTML]{DDDDDD}-- & \cellcolor[rgb]{0.573,0.757,0.617}77.04 & \cellcolor[rgb]{0.489,0.727,0.545}85.04 & 2.13 $\pm$ 0.07 \\
Gaussian Grouping & \cellcolor[rgb]{0.819,0.395,0.470}15.81 & \cellcolor[rgb]{0.828,0.492,0.552}22.96 & \cellcolor[HTML]{DDDDDD}-- & \cellcolor[rgb]{0.678,0.795,0.705}67.17 & 2.94 $\pm$ 0.08 \\
InFusion & \cellcolor[rgb]{0.808,0.275,0.370}\color{black}6.99 & \cellcolor[rgb]{0.818,0.383,0.461}14.96 & \cellcolor[rgb]{0.839,0.627,0.664}32.83 & \cellcolor[HTML]{DDDDDD}-- & 3.45 $\pm$ 0.08 \\
\bottomrule
\end{tabular}
}
\end{wraptable}

% \begin{table}
%     \caption{
%         User study. Each cell contains the tie-adjusted win-rate in \%, row-vs-column.
%         }
%     \label{tab:user-study-global}
%     \vspace{-2mm}
%     \centering
%     \small
%     %\renewcommand{\arraystretch}{1.2}
% \begin{tabular}{lccccc}
% \toprule
% Method & Ours & Inpaint360GS & Gaussian Grouping & InFusion & Expected rank $\downarrow$ \\
% \midrule
% Ours & \cellcolor[HTML]{DDDDDD}-- & \cellcolor[rgb]{0.601,0.767,0.640}74.48 & \cellcolor[rgb]{0.504,0.732,0.558}83.59 & \cellcolor[rgb]{0.392,0.692,0.463}94.16 & \textbf{1.48 $\pm$ 0.08} \\
% Inpaint360GS & \cellcolor[rgb]{0.831,0.527,0.581}25.52 & \cellcolor[HTML]{DDDDDD}-- & \cellcolor[rgb]{0.569,0.756,0.613}77.44 & \cellcolor[rgb]{0.494,0.729,0.549}84.52 & 2.13 $\pm$ 0.08 \\
% Gaussian Grouping & \cellcolor[rgb]{0.820,0.403,0.477}16.41 & \cellcolor[rgb]{0.827,0.487,0.547}22.56 & \cellcolor[HTML]{DDDDDD}-- & \cellcolor[rgb]{0.681,0.796,0.708}66.92 & 2.94 $\pm$ 0.09 \\
% InFusion & \cellcolor[rgb]{0.807,0.259,0.357}\color{white}5.84 & \cellcolor[rgb]{0.819,0.391,0.466}15.48 & \cellcolor[rgb]{0.840,0.630,0.667}33.08 & \cellcolor[HTML]{DDDDDD}-- & 3.45 $\pm$ 0.09 \\
% \bottomrule
% \end{tabular}
% \end{table}

\pgfmathsetmacro{\nvotes}{1401}
\pgfmathsetmacro{\nplayers}{29}
\pgfmathsetmacro{\votespplayer}{\nvotes / \nplayers }
\textbf{User study.} As detailed in Appendix~\ref{appendix:human_metric}, automated metrics poorly correlate with human preference in our inpainting setting: we thus further assess the visual quality of our approach via a blind user study.
Across the 29 scenes from the COR-NeRF, Inpaint360 and 360-USID datasets, we randomly sample a viewpoint from a held-out set, and render an input view (with the object present) and the inpainted results of two methods, displayed in random left-right order.
Users are asked which result looks better, with the options of skipping (e.g., if the region is occluded from this view) or signaling a tie, in case both results look equally good or bad.
%Each user is asked to rate at least 50 image pairs. 
During the study, \nplayers~participants rated a total of \nvotes~pairs, an average of \pgfmathprintnumber[fixed, precision=1]{\votespplayer} per user. 
\Cref{tab:user-study-global} shows that our method is overwhelmingly preferred by human evaluators, achieving a 75\% preference rate against the strongest baseline and attaining an expected rank of $1.48$.

\begin{table*}[t]
    \caption{
        Averaged metrics across datasets.
        Bold and underline indicate best- and second-best.
    }
    \label{tab:benchmark-aggregate}
    \vspace{-2mm}
    \centering
    \small
    \resizebox{\linewidth}{!}{%
    \begin{tabular}{lcccccc}
        \toprule
        \textbf{Method} & \textbf{LPIPS} $\downarrow$ & \textbf{FID} $\downarrow$ & \textbf{Local-FID} $\downarrow$ & \textbf{KID} $(\times 10^3)\downarrow$ & \textbf{Human Pref.} $(\%) \uparrow$ & \textbf{Runtime (s)} $\downarrow$ \\
        \midrule
        Gauss. Grouping & $0.255$ & $66.98$ & $66.53$ & $18.02$ & $34.8$ & $780.0$ \\
        InFusion & $0.250$ & $88.71$ & $87.74$ & $40.31$ & $18.6$ & $212.9$ \\
        Inpaint360GS & $\underline{0.236}$ & $\underline{51.17}$ & $\underline{52.43}$ & $\underline{6.16}$ & $\underline{63.6}$ & $\underline{202.1}$ \\
        Ours & $\bm{0.235}$ & $\bm{49.96}$ & $\bm{49.21}$ & $\bm{6.08}$ & $\bm{82.9}$ & $\bm{176.8}$ \\
        \bottomrule
    \end{tabular}
    }%
\end{table*}

\begin{figure}
    \centering
    \includegraphics{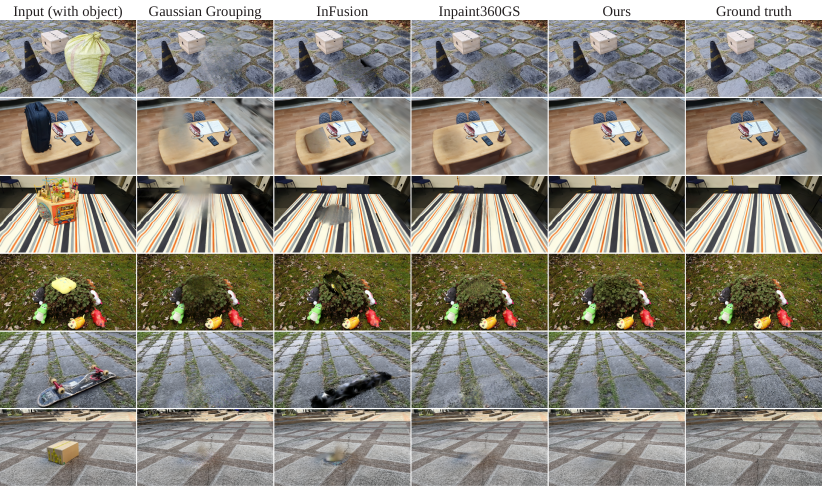}
    \vspace{-6mm}
    \caption{Qualitative comparison on the COR-NeRF, Inpaint360 \& 360-USID datasets (top to bottom, two rows each, best viewed zoomed in). See the 
    %appendix and electronic supplemental for more results and a 
    supplemental video to assess multiview-consistency. %\MF{TODO: replace carton, we already show that for the ablation}
    }
    \label{fig:qualitative_comparison}
\end{figure}

\textbf{Qualitative results.}
Qualitative inspection of our results confirms the quantitative analysis: our method consistently achieves highly plausible and sharp inpainting results that respect well the visual structure of the scene and integrate seamlessly with their surroundings.
Other methods often produce blurry and fuzzy regions or show significant color shifts or artifacts. 
We show a subset of our results in \Cref{fig:qualitative_comparison}; for a full gallery and a comparison against baselines on all scenes, please see Appendix~\ref{appendix:subsec:per-scene-visualization}.% and the electronic supplemental. %(removed for preprint)

\subsection{Ablation study}

\textbf{Attention masking.} This ablation uses the same encoded context \ac{SLAT}s $\z_c$ and predicted structure $O$ but removes the attention masking mechanism (see \Cref{fig:attention-masking}).
The structured latent conditional sampling stage hence decays back to the original conditional sampling of $\mathcal G_L$ used by RePaint, as implemented by \citet{xiang_structured_2025a}. 
The number of integration steps, backwards jump length and number of jumps is kept identical for both. 

\textbf{Structure completion.} We ablate our structure completion U-Net and
%use the same encoded context latents $\z_c$ as the original run. 
instead use RePaint sampling of the sparse structure model $\mathcal G_S$ in a way identical to \citet{xiang_structured_2025a}, with 50 integration time steps. 
After obtaining the completed structure $O$, we perform attention-masked \ac{SLAT} generation with the usual hyperparameters.

\textbf{Decoder adaptation.} For this ablation, we predict structure using our U-Net and run our full latent pipeline (with progressive context conditioning and self-attention masking), and simply refrain from applying the scene-conditioned parameters $\phi^*$ to $\mathcal{D}_\theta$ during the final decoding of generated latents.

\begin{figure}
    \centering
    \includegraphics{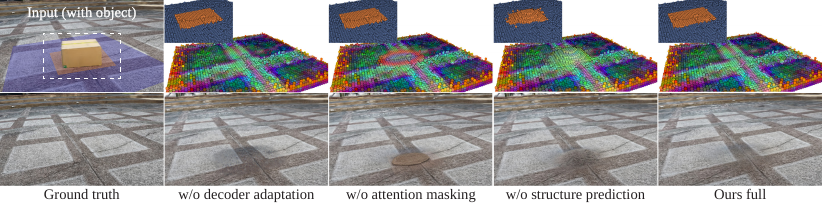}
    \vspace{-6mm}
    \caption{Ablation visualization. We show the context and mask regions as overlays on the input image. 
    Each column shows the completed structure (inset, top row), the PCA of the inpainted features after the integration process (principal three components in RGB, middle row), and the final decoded result in the bottom row.
    }
    \label{fig:ablations}
\end{figure}

\begin{wraptable}[8]{r}{0.67\linewidth}
\vspace{-\intextsep}
    \centering
    \small
    \caption{
        Perceptual metrics for our method and its ablations.%, with key components removed.
        %\NL{Re-gen results for KID decimals}
    }
    \label{tab:ablations}
    \vspace{-2mm}

    \setlength{\tabcolsep}{3pt}
    \resizebox{\linewidth}{!}{%
    \begin{tabular}{lcccc}
\toprule
Method & LPIPS $\downarrow$ & FID $\downarrow$ & Local-FID $\downarrow$ & KID ($\times 10^3$) $\downarrow$ \\
\midrule
Ours & $\bm{0.235}$ & $\bm{49.961}$ & $\bm{49.214}$ & $\bm{6.080}$ \\
Ours (w/o attention masking) & $\underline{0.236}$ & $\underline{52.806}$ & $\underline{51.718}$ & $\underline{9.497}$ \\
Ours (w/o axis alignment) & $0.241$ & $55.336$ & $54.661$ & $10.030$ \\
Ours (w/o structure prediction model) & $0.238$ & $57.260$ & $59.181$ & $11.893$ \\
Ours (w/o decoder adaptation) & $0.250$ & $68.776$ & $63.755$ & $24.867$ \\

\bottomrule
    \end{tabular}}
\end{wraptable}
\Cref{tab:ablations} reports perceptual metrics for all ablations. 
Decoder adaptation has the largest impact on perceptual quality, with its removal substantially degrading all four metrics, while the structure completion model provides the second-largest contribution, particularly for Local-FID.
Attention masking yields a smaller albeit consistent improvement, but contributes significantly to seamless blending and overall visual quality, as the qualitative visualizations in \Cref{fig:ablations} show. 

% \begin{figure}
%     \centering
%     \includegraphics{figures/ablations/ablation_figure_raster.ai}
%     \caption{Qualitative ablation study on a representative scene}
%     \label{fig:ablation_figure }
% \end{figure}

% \begin{table}
%     \caption{
%         Perceptual metrics for our method, with key components removed. \NL{Will regenerate the results to get back the decimals from the KID metric since we need to multiply it by 1000}}
%     \label{tab:ablations}
%     \vspace{-2mm}
%     \centering
%     \small
% \begin{tabular}{lcccc}
% \toprule
% Method & LPIPS $\downarrow$ & FID $\downarrow$ & Local-FID $\downarrow$ & KID ($\times 10^3$) $\downarrow$\\
% \midrule
% Ours & $\bm{0.235}$ & $\bm{49.775}$ & $\bm{49.162}$ & $\bm{5.788}$ \\
% Ours (w/o attention masking) & $\underline{0.236}$ & \underline{$52.806$} & \underline{$51.718$} & \underline{$9.595$} \\
% Ours (w/o structure completion model) & $0.238$ & $57.260$ & $59.181$ & $12.051$ \\
% Ours (w/o decoder adaptation) & $0.250$ & $68.776$ & $63.755$ & $24.814$ \\

\begin{figure}
    \centering
    \includegraphics{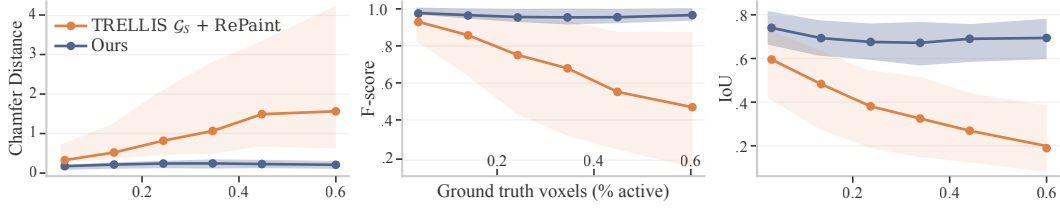}
    \vspace{-6mm}
    \caption{Sparse structure reconstruction metrics against masked-to-context voxel ratio. Our proposed structure completion network strongly outperforms TRELLIS $\mathcal{G}_S$ + RePaint for all context ratios.}
    \label{fig:repaint-vs-unet}
\end{figure}

% \bottomrule
% \end{tabular}
% \end{table}
We further ablate our structure completion model against the baseline proposed by TRELLIS \citep{xiang_structured_2025a}, applying the RePaint algorithm \citep{lugmayr_repaint_2022a} to the sparse structure flow model $\mathcal{G}_S$. 
We evaluate both approaches on a held-out dataset of 3DGS scenes unused during training of our structure completion U-Net, and report their relative performance in \Cref{fig:repaint-vs-unet}. 
We find the sparse structure completion of the TRELLIS baseline to strongly degrade as the masked region size increases compared to context, while our structure completion model offers both accurate and consistent predictions even at high masked voxel ratios.

%%%%%%%%%%%%%%%%%%%%%%%%%%%%%%%%%%%%%%%%%%%%%%%
\section{Discussion \& Conclusion}

%\elie{NB: FID is the best metric, but it remains a poor proxy of inpainting quality. -> See bush example. Also show examples where user study != metric.} \MF{maybe in suppl? I wouldnt stress it too much here because we win in both user study and metrics, so we'd be undermining our own wins a bit}

\textbf{Limitations.} 
While our method successfully inpaints 3DGS assets in unbounded 360° scenarios and outperforms other current approaches in both perceptual metrics and human preference, it is not free of limitations, which we highlight here as avenues for future work. 
First, we observe that content generation of complex and intricate patterns in latent space is biased towards axis-aligned patterns, and struggles with misaligned motifs. 
We devise an axis-alignment heuristic to overcome this, but future research could help mitigate this at a structural level, e.g., inspired by SO(3)-equivariant structure alignment mechanisms~\citep{ding_scalable_2025} or by leveraging recent approaches to equivariant 3D-native priors~\citep{ji_symtrellis_2026}.
Moreover, due to architectural limitations of the TRELLIS VAE decoder $\mathcal{D}_\theta$, our approach does not generate the higher-order spherical harmonic components of the inpainted splats. 
While we plan to investigate this in future work, our core framework is agnostic to the specific choice of backbone, and thus could accommodate future priors supporting higher-order spherical harmonics.

Additionally, despite operating with a lower runtime than existing 2D-based approaches, our method requires around three minutes for inpainting. 
As \Cref{tab:runtime} shows, a large fraction of that time is allocated to the decoder adaptation stage. Rather than training a single adapter per scene, future work will focus on training a single multi-scene adapter, which will amortize the training cost over successive interactive edits of the same scene. 
%\MF{tmp rm'ed NL's comment about disjoint bboxes}

%\NL{Not sure we need to talk about this if we're short in space. It's not a big deal. Our approach fits the 3D mask in a bounding box in order to create the target and context regions and thus processes disjoint masks in a unique, larger bounding box. While achieving good results on an evaluation scene containing disjoint masks, an alternative solution could be to separately process connected mask components one at a time.}

Finally, pixel-reference metrics such as PSNR and SSIM are ill-suited for evaluating the output of a stochastic process such as diffusion-based inpainting, since perceptually plausible variations (e.g., a different realization of the same wood grain) may incur a large pixel-wise error despite being visually convincing (see our discussion in \Cref{sec:results} and Appendix~\ref{appendix:subsec_additional-results}). Extending perceptually motivated, inpainting-specific quality metrics \citep{chen_assessing_2024a} to 3DGS inpainting is beyond the scope of this work, but remains an important direction for future research.

\textbf{Conclusion.} 
We propose a novel, 3D-native framework for inpainting regions of real-world, unbounded Gaussian splatting scenes. 
Our approach leverages a powerful generative prior to conditionally create plausible content in latent space. 
We introduce three separate mechanisms improving the quality and faithfulness of inpainting: a learned structure completion network, a self-attention masking strategy to improve information exchange between context and generated latents, and a scene-specific decoder adaptation module. 
Our method completes in a few minutes, outperforms existing 3DGS inpainting methods, achieves the best aggregate metrics on a variety of scenes from three different datasets, and is consistently ranked highest by human preference.
To the best of our knowledge, our method is the first 3DGS inpainting approach to perform content generation in the representation space of a 3D-native prior, opening important future perspectives for interactive 3D scene interaction and editing.
\bibliography{biblio}
\bibliographystyle{iclr2027_conference}

%%%%%%%%%%%%%%%%%%%%%%%%%%%%%%%%%%%%%%%%%%%%%%%
\newpage
\appendix

\FloatBarrier

\begin{center}
  {\LARGE Appendix}
\end{center}

%We propose here an appendix to the present work. In addition, supplementary material is provided as a compressed zip file. We encourage the reader to view the two short videos demonstrating our framework, as well as the electronic gallery in \texttt{electronic\_supplemental/index.html}.%(removed for preprint)

\section{Additional results}
\label{appendix:subsec_additional-results}

\subsection{Reference-based metrics}
\label{appendix:human_metric}

% \begin{wrapfigure}{r}{0.48\linewidth}
%     \vspace{-\intextsep}
%     \centering
%     \includegraphics{figures/supplementary/metrics_are_bad.pdf}
%     \vspace{-6mm}
%     \caption{Inpainting results on the Redbull scene from the Inpaint360 dataset. Even though our method produces less artifacts and a visually more coherent continuation of the striped surface, Inpaint360GS is achieves lower FID and KID scores.}
%     \label{fig:metrics_are_bad}
% \end{wrapfigure}

\begin{figure}[b!]
    \centering
    \includegraphics{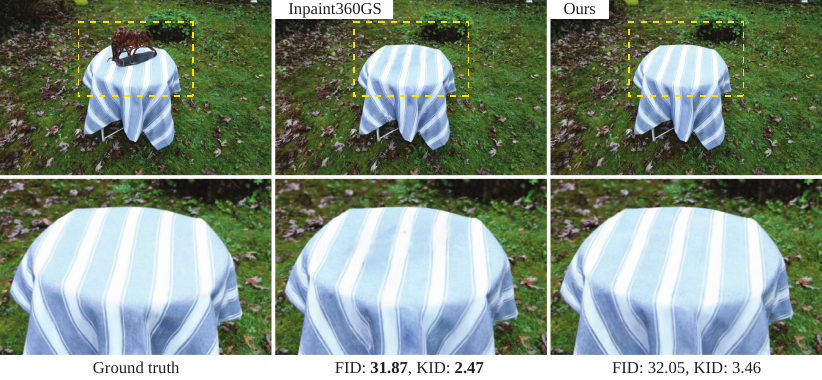}
    \vspace{-6mm}
    \caption{
    Inpainting results on the ``Red bull'' scene from the Inpaint360 dataset. Even though Inpaint360GS achieves better (lower) FID and KID scores, our method produces a more coherent continuation of the striped surface, and is preferred by 91.7\% of human evaluators.
    }
    \label{fig:metrics_are_bad}
\end{figure}

\begin{figure}[b!]
    \centering
    \includegraphics{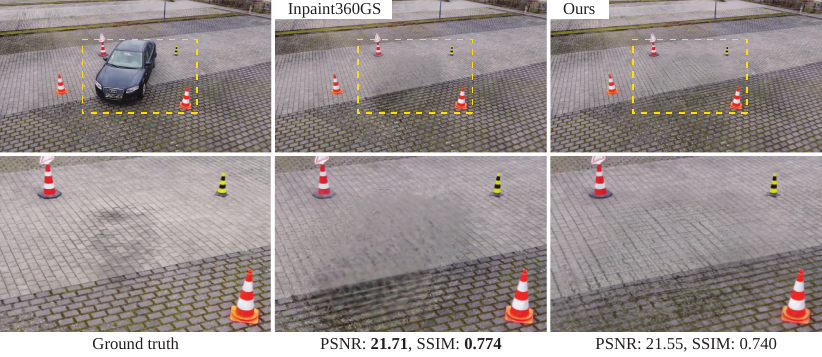}
    \vspace{-6mm}
    \caption{Inpainting results on the ``Car'' scene from the Inpaint360 dataset. Similar to \Cref{fig:metrics_are_bad}, Inpaint360GS performs better on the pixel-aligned metrics PSNR and SSIM, even though our method produces a more plausible and notably sharper inpainting result, making it the preferred choice for 100\% of human evaluators.} 
    \label{fig:metrics_are_bad_psnr}
\end{figure}

\begin{figure}[t!]
    \centering
    \includegraphics{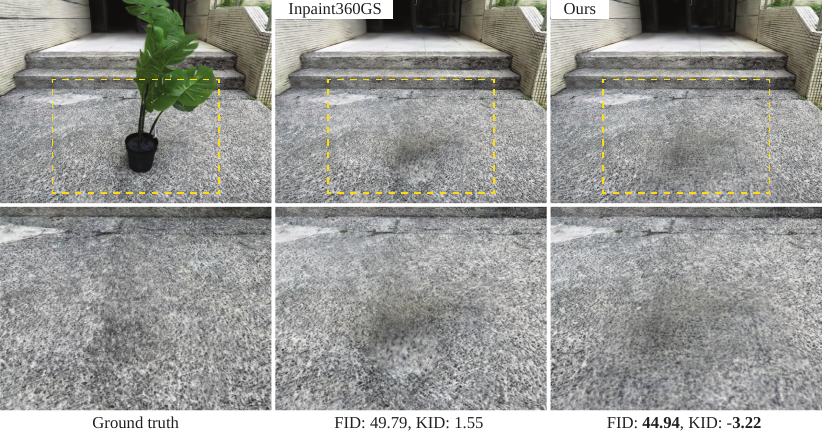}
    \vspace{-6mm}
    \caption{Inpainting results on the ``Plant'' scene from the 360-USID dataset. Here, Ours performs better on the distributional FID and KID metrics, despite being preferred by only 37.5\% of human evaluators compared to Inpaint360GS.} 
    \label{fig:metrics_are_bad_plant}
\end{figure}

In addition to the results from \Cref{tab:benchmark-aggregate}, we report the reference-based distortion metrics PSNR and SSIM and the reference-based perceptual metric LPIPS in \Cref{tab:fidelity-aggregate}. Although very widely used in 3DGS inpainting benchmarks~\citep{wu_aurafusion360_2025a, lee_gpgs_2026, chen_gsroadpatching_2025}, we argue that PSNR and SSIM are not satisfactory evaluators of inpainting quality. Indeed, these metrics report fidelity to a reference, which is in the present case a different 3DGS scene with the object removed. This is the source of many possible inconsistencies stemming from light or environment change. Furthermore, these metrics only evaluate the fidelity of inpainting with respect to one ground truth, not accounting for the large number of valid inpainting solutions admitted in the target region. Despite outperforming existing methods on PSNR and being strongly competitive on SSIM, we prefer to rely on distributional perceptual metrics such as the Fréchet Inception Distance (FID, \cite{heusel_gans_2017}) or the Kernel Inception Distance (KID, \cite{binkowski_demystifying_2018}).

%\begin{table}
\begin{wraptable}[11]{r}
{0.58\linewidth}
\vspace{-\intextsep}
    \caption{
        Aggregate image-fidelity metrics, averaged across all datasets.
        Bold and underline indicate the best and second-best result in
        each column, respectively.
    }
    \label{tab:fidelity-aggregate}
    \centering
    \small
    \setlength{\tabcolsep}{8pt}
    \begin{tabular}{lccc}
        \toprule
        \textbf{Method} & \textbf{PSNR} $\uparrow$ & \textbf{SSIM} $\uparrow$ & \textbf{LPIPS} $\downarrow$ \\
        \midrule
        Gaussian Grouping & $20.58$ & $0.617$ & $0.255$ \\
        InFusion & $20.68$ & $0.628$ & $0.250$ \\
        Inpaint360GS & $\underline{21.94}$ & $\bm{0.644}$ & $\underline{0.236}$ \\
        Ours & $\bm{21.95}$ & $\underline{0.640}$ & $\bm{0.235}$ \\
        \bottomrule
    \end{tabular}
\end{wraptable}
We observe that FID and its full-resolution variant Local-FID are the second-best correlated metrics with human preference, behind the reference-based LPIPS metric (Appendix~\ref{appendix:human_metric}), while KID only weakly correlates with human judgment. \Cref{tab:metric-human-agreement} proposes a comparative alignment analysis of human preference and quantitative metrics, computed only for Ours-Inpaint360GS duels. The table cells illustrate the relative metric superiority of Ours against Inpaint360GS, colored in green when our method performs better and red when it performs worse. The color shade is proportional to the log-ratio between the two metrics, except for PSNR and the two KID metrics, where we use the difference, as PSNR is already a logarithmic metric and KID can take negative values.
Remarkably, even the best-correlated metrics (LPIPS, FID and Local-FID) exhibit a timid Spearman $\rho$ correlation coefficient with human preference of respectively 0.48, 0.40 and 0.40, indicating only partial agreement with human preference. In contrast, reference-based distortion metrics such as PSNR and SSIM are expectedly poor predictors of human preference, with Spearman $\rho$ coefficients of 0.14 and 0.21 only. We illustrate these phenomena with qualitative visualizations: \Cref{fig:metrics_are_bad,fig:metrics_are_bad_psnr} show cases where Inpaint360GS surpasses \textsc{Wings} on metrics despite our method being preferred by humans, and \Cref{fig:metrics_are_bad_plant} shows a case where humans prefer Inpaint360GS but our result obtains better metrics. 

In addition to justifying our choice to omit PSNR and SSIM from our quantitative evaluation protocol, these results highlight the lack of 3D-specific metrics able to assess inpainting quality in a way that aligns with human judgment.

\subsection{User study} \label{appendix:user-study}

The user study was conducted through an interactive web application, displaying a render of the input scene, containing an object to be removed, and renders of two randomly chosen methods between Gaussian Grouping, InFusion, Inpaint360GS and \textsc{Wings}. The opposing methods are sampled uniformly: in the final study, all method pairs received an average of 233.5 head-to-head comparisons. In particular, Ours and Inpaint360GS were involved in 231 head-to-head comparisons, distributed among \nplayers~participants. In order to avoid bias due to position, the left-right ordering of the methods is chosen at random. Among the \nvotes~ratings, 111 exact $\text{(scene, view, method pair)}$ tuples were shown to at least two different users. In these situations, the human participants agreed on identical ratings in $77.1\%$ of observations, with a bootstrap 95\% confidence interval of $[69.7\%, 84.6\%]$. The Krippendorff's $\alpha$ inter-rater reliability computed over these identical tuples is $0.680$, with a per-duplicate bootstrap 95\% confidence interval of $[0.579, 0.776]$.
Under the Davidson model, 26 out of \nplayers~participants obtain the same complete ordering of the four methods. 

The human-preference score reported in \Cref{tab:benchmark-aggregate} is computed by fitting a Bradley-Terry-Davidson model~\citep{bradley_rank_1952, davidson_extending_1977} to the pairwise preferences registered during the user study. We report the expected tie-adjusted score against an average-skill opponent under this model: 
\begin{equation*}
    \text{preference}_i = \mathbb{P}(i \succ \mathrm{avg})
        + \frac{1}{2}
        \mathbb{P}(i \sim \mathrm{avg}).
\end{equation*}
Under the same model, the expected rank reported in \Cref{tab:user-study-global} is computed as follows:
\begin{equation*}
    \text{Expected rank} =N_{\mathrm{methods}}
        -
        \sum_{j \neq i}
        \left(
            \mathbb{P}(i \succ j)
            + \frac{1}{2}
            \mathbb{P}(i \sim j)
        \right).
\end{equation*}
Finally, the human preference score of Ours against Inpaint360GS reported in \Cref{tab:metric-human-agreement} is simply computed using a tie-adjusted preference with two players: 
$$
\text{preference}_i = \frac{W_{ij} + \frac{1}{2}T_{ij}}{N_{ij}},
$$

\subsection{Runtime analysis}

\begin{wraptable}[12]{r}
{0.5\linewidth}
\vspace{-\intextsep}
    \caption{
        Per-stage runtime of our method.
    }
    \label{tab:runtime}
    \vspace{-2mm}
    \centering
    \small
    \begin{tabular}{lcc} 
        \toprule
        Stage & Ours & Ours (XS) \\
        \midrule

        Context encoding & 20.40 & 20.13 \\
        Decoder adaptation & 103.02 & 23.58 \\
        Structure prediction & 0.56 & 0.56 \\
        Conditional latent sampling & 51.20 & 29.59 \\
        Decoding and merging & 1.66 & 1.67 \\
        \midrule
        \textbf{Total runtime (s) $\downarrow$} & 176.84 & \textbf{75.39} \\
        \textbf{FID $\downarrow$} & \textbf{49.96} & 57.14 \\
        \bottomrule
    \end{tabular}
\end{wraptable}
We describe here in detail the runtime of our method and make its per-stage complexity explicit. Our framework comprises five main stages: \textbf{(a)} context encoding; \textbf{(b)} scene-specific decoder adaptation; \textbf{(c)} sparse structure completion; \textbf{(d)} conditional latent sampling and \textbf{(e)} final decoding and merging. 
Among these, stages (b) and (d) make up the overwhelming majority of the total inpainting time: the exact average runtimes of each stages are detailed in \Cref{tab:runtime}. Even though our method, performing inpainting in under 3 minutes, is suitable for interactive usage, we also benchmark a stripped-down lightweight version (Ours-XS), where we fine-tune the decoder for 100 steps only with a larger learning rate of $5 \times 10^{-4}$ and a rank-8 adapter, and generate latent features for the inpainted region with fewer integration steps and a reduced RePaint schedule (50 steps, with 3 resamples and a jump size of 5). This reduced computational complexity allows us to perform inpainting in around 75 seconds with only a limited perceptual fidelity drop: on the three benchmark datasets, we achieve an FID of $57.14$ and a LPIPS of $0.2366$, comparable with the existing 2D-based approach of Gaussian Grouping~\citep{ye_gaussian_2024a} at a fraction of the runtime. This proves the ability of our lightweight framework to operate reliably at a reduced runtime, an important step towards real-time radiance field editing applications.

\section{Implementation details}

\subsection{Implementation details of our method}
\label{appendix:subsec:implem}

\paragraph{Generative prior.} We use the publicly released TRELLIS model as the generative prior for all experiments in this work. In particular, we use the pre-trained \texttt{TRELLIS-image-large} weights for the SLAT VAE encoder and decoder and for the SLAT flow transformer $\mathcal{G}_L$ from commit \texttt{442aa1e} (Nov 5, 2025). For ablation studies implementing the sparse structure flow model, the model weights are also those from \texttt{TRELLIS-image-large}.

\paragraph{Context encoding.} As described in \Cref{subsec:encoding_and_structure_completion}, we encode the original scene to \ac{SLAT} by selecting a region lying inside a bounding box. The generative prior used in this work, TRELLIS \citep{xiang_structured_2025a}, operates in a latent space with a spatial resolution of $64$.  However, 3DGS scenes assume widely varying extents, and therefore large scenes cannot always be discretized with minimal information loss in a $64^3$ space. The size of the context bounding box must be small enough to faithfully encode the appearance of the content, but large enough to provide sufficient information for sampling from the context-conditional distribution $\pgl(\z_m | z_c)$. Although different scenes may naturally admit varying optimal context bounding box sizes, we fix the size of the bounding box in the experiments of this paper to the smallest bounding box where, after voxelization based on Gaussian primitive centers, the ratio of context to target region voxels is greater than 6. The orientation of this bounding box is detailed in Appendix~\ref{appendix:subsec:autoalign} below. The bounding box, lying in world space, is then normalized to the unit cube $[-0.5, 0.5]^3$, and Gaussian center positions are discretized to voxels lying in a $64^3$ grid, which is the spatial resolution of the TRELLIS space. The normalization maps the $\mathbf{c}_\text{world} = (x, y, z)$ coordinates to:
$$\mathbf{c}_{\text{slat}} = \frac{\mathbf{c}_\text{world} - \mu}{s}, $$
where $\mu$ is the position of the bounding box center and $s$ is the size of the largest bounding box axis.

We then compute the DINOv2~\citep{oquab_dinov2_2023} features of 36 views sampled from a virtual orbit rotating around the normalized context, determined in the same way as \cite{xiang_structured_2025a}, with a render size of $518 \times 518$. As 3DGS scenes can contain large, low-opacity unwanted floater artifacts, we additionally drop voxels that are isolated in a $3\times3 \times 3$ neighborhood.

After generating the masked region Gaussians $G_m \sim \pgl(G_m | G_c)$, we unnormalize their positions $\mathbf{c}_\text{slat}$ back to world space in order to merge generated and context Gaussians: $\mathbf{c}_\text{world} = \mathbf{c}_\text{slat}\cdot s + \mu$. Additionally, we apply a linear opacity ramp to the generated Gaussians lying up to 5 voxels outside of the target region's boundary, in order to ensure seamless transition between $G_c$ and $G_m$.

\paragraph{Sparse structure prediction.} Our structure prediction network implements a 3D residual U-Net architecture with attention in the bottleneck. We instantiate the model with an initial 64 convolutional channels, doubling over each of the three downsampling stages to reach 512 channels in the bottleneck. This results in a model with 49.7M parameters. The training protocol of our proposed sparse structure prediction model is described in \Cref{appendix:subcsec:structure-dataset}.

\paragraph{Progressive context conditioning.}
When evaluating the re-noised known context latents $\hat\z$ in \Cref{eq:repaint}, we slightly remap the noise level, using $\left(10^{-5} + (1 - 10^{-5}) t\right) \epsilon$ rather than $t\epsilon$. This exactly follows the protocol established by TRELLIS~\citep{xiang_structured_2025a}. We apply the RePaint scheme using 100 integration steps and 5 backwards jumps of length 10.

\paragraph{Self-attention masking.}
 We conduct a hyperparameter grid search to determine the optimal flow transformer layers and integration time steps on which to perform self-attention masking. After a first coarse search, we observe that masking the late layers of the model at high-noise integration time steps is the best performing strategy. We then compare masking layers in range $19\!-\!24$, $18\!-\!23$ and $19\!-\!23$ while varying from $t=1.0$ and $t=t_\text{stop}$ with $t_\text{stop} \in \{0.55, 0.6, 0.65, 0.7, 0.75, 0.8\}$. Hence, we choose to apply self-attention masking to layers $18\!-\!23$ of the 24-layer $\mathcal{G}_L$ flow model between $t=1.0$ and $t=0.8$. Applying self-attention masking at high-noise time steps allows the flow model to reconcile the trajectories of context and masked latents while the coarse appearance of the structured latents is forming, while unrestricted self-attention at later time steps allows target latents to attend to themselves in order to harmonize fine-grained detail.

\needspace{0.5\textheight}
\begin{wrapfigure}[14]{r}{0.5\linewidth}
    \vspace{-\intextsep}
    \centering
    \includegraphics[width=\linewidth]{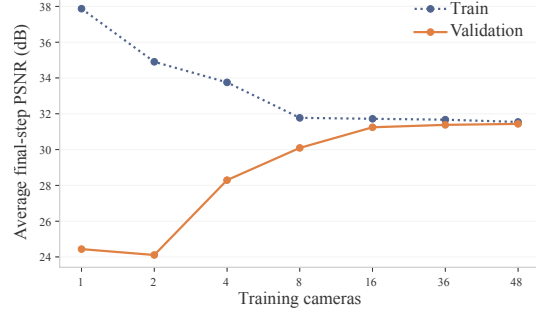}
    \vspace{-6mm}
    \caption{Train and validation viewpoint PSNR during scene-conditioned decoder adaptation vs.\ number of training cameras.}
    \label{fig:lora_cameras}
\end{wrapfigure}
 \paragraph{Scene-conditioned decoder adaptation.}
 Our scene-conditioned decoder adaptation module consists in a rank-16 LoRA adapter~\citep{hu_lora_2021a} attached to every block of the VAE decoder. Adapters are attached to each block's attention and MLP layers. Additionally, we unfreeze the final linear layer of the decoder, which is responsible for predicting the attributes of the 32 Gaussian primitives predicted for each voxel.

 We train the adapter using the AdamW~\citep{loshchilov_decoupled_2018} optimizer with a learning rate of $10^{-4}$ for 600 steps. The supervisory renders are obtained by sampling 36 views from a virtual trajectory covering a 360° orbit around the context region, inherited from the virtual trajectory protocol of~\citep{xiang_structured_2025a}, which we modify by applying a sinusoidal modulation to the orbit radius, in order to capture range-dependent variations in the appearance of Gaussians. We additionally hold out 6 viewpoints from this trajectory to evaluate unseen viewpoint generalization.
 
 The photometric loss $\mathcal{L}_{\text{data}}$ used to optimize the parameters $\phi$ is:
\begin{equation}
    \mathcal{L}_\text{data} = \mathcal{L}_1 + \lambda_\text{D-SSIM} \mathcal{L}_\text{D-SSIM} + \lambda_\text{LPIPS} \mathcal{L}_\text{LPIPS}
\end{equation}
We use $\lambda_\text{D-SSIM} = 0.2$ and $\lambda_\text{LPIPS} = 0.2$ for all experiments in this paper.

% \begin{figure} %TODO maybe put this in wrapfig
%     \centering
%     \includegraphics[width=0.5\linewidth]{figures/supplementary/lora_cameras.ai}
%     \caption{Train and validation viewpoint PSNR during scene-conditioned decoder adaptation vs. number of training cameras.}
%     \label{fig:lora_cameras}
% \end{figure}
\Cref{fig:lora_cameras} illustrates the impact of changing the number of training cameras for the scene-conditioned decoder adaptation stage. While using 1 to 8 cameras does not result in satisfying reconstruction, 16 or more cameras allow the adapter to faithfully match the original scene's appearance, with diminishing returns as the number of training cameras increase.

\subsection{Baseline implementation details}

We benchmark existing 3D Gaussian splatting methods in a controlled environment. All methods, including ours, receive an identical 3DGS scene as input as well as a predefined 3D segmentation mask. The per-Gaussian segmentation mask is created in 3D editing software and encompasses all Gaussians corresponding to the object to remove. The use of an existing 3D mask allows us to bypass the expensive and error-prone Gaussian segmentation stages of existing pipelines, enabling direct comparison of the inpainting mechanism of each.

\paragraph{Gaussian Grouping.} We follow the official implementation of Gaussian Grouping~\citep{ye_gaussian_2024a}. Instead of optimizing the 3D Gaussian splatting scene from scratch and performing the Gaussian segmentation pipeline, we use the same 3D mask $\mathcal{M}$ used by our method (and other implemented baselines), allowing for fair comparison of the inpainting frameworks themselves. 

\paragraph{Inpaint360GS.}
Our experimental protocol, comprising an existing 3D mask, bypasses the object association and ID distillation stages of Inpaint360GS, instead using a known correct 3D segmentation mask identical to other approaches for fair comparison. We otherwise use the authors' released inpainting pipeline as provided, including virtual view generation, conditional color and depth inpainting, color-depth fusion and 3DGS fine-tuning. As described in the authors' implementation details, we run the inpainting pipeline with $2000$ fine-tuning steps and loss parameters $\lambda_{D-SSIM}=0.2$ and $\lambda_{LPIPS}=0.005$. Whilst the authors prescribe running the method at $1/4$ resolution, we found that to have adverse effects on datasets where reference images have lower resolution. Instead, we evaluated the render size of the LaMa inpainted views and of the subsequent optimization renders at $r=\{256, 512, 1024, 2048\}$ (clamped at the maximum image resolution), and found a render size of 512 to offer the best results (On the COR-NeRF dataset, $r=512$ obtains an FID of $36.10$ while $r=256, 1024$ and $2048$ attain respectively $44.1, 39.6$ and $39.3$). 
Additionally, we were unable to observe the reported 3 minutes inpainting time, instead measuring a slightly higher inpainting time of 202 seconds. Since our input modality does not include per-Gaussian instance identities for non-target objects, we do not use Inpaint360GS's auxiliary object removal mechanism in the inpainting pipeline, rather evaluating the inpainting and refinement component of the method.

\paragraph{InFusion.} We implement InFusion~\citep{liu_infusion_2024a} following the official code provided by the authors. We notice that InFusion struggles with 360° unbounded scene inpainting, offering poor depth unprojection and overall inpainting quality on our benchmarked datasets. As the official protocol does not restrict the position of the generated Gaussians, this resulted in large artifacts scattered in the direction of the unprojected depth with respect to the chosen view. Therefore, we decided to discard the generated Gaussians lying outside of the 3D mask $\mathcal{M}$, expanded by 5\%, in order to provide a more faithful comparison of target region inpainting quality.

\FloatBarrier
\subsection{Automatic orientation of the \ac{SLAT} grid}
\label{appendix:subsec:autoalign}

\begin{figure}[t]
    \centering
    \includegraphics{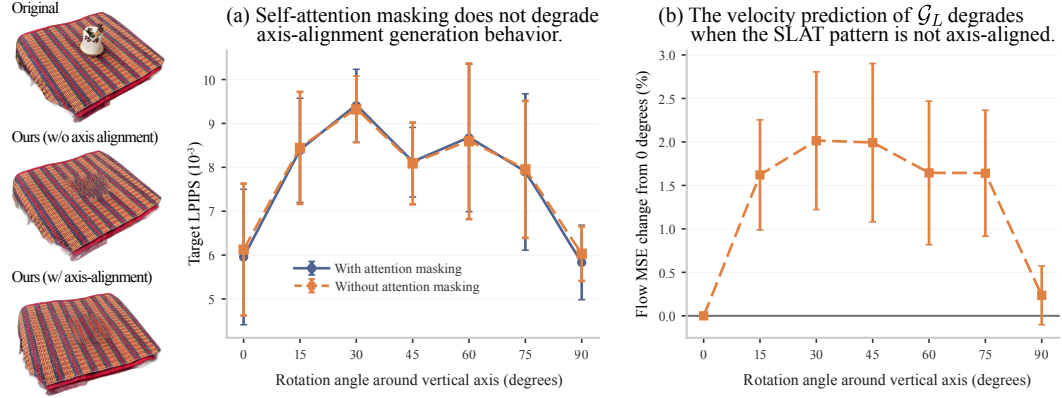}
    \caption{Illustration of the generative prior's bias towards generating axis-aligned patterns. (a) While self-attention masking neither degrades nor improves perceptual loss during inpainting for these examples, inpainting perceptually worsens between 0 and 90° rotations. (b) The flow velocity mean squared error also increases as the patterned SLAT grids are rotated, compared to an axis-aligned velocity at 0°.}
    \label{fig:axis_bias}
\end{figure}

\begin{figure}
    \centering
    \includegraphics{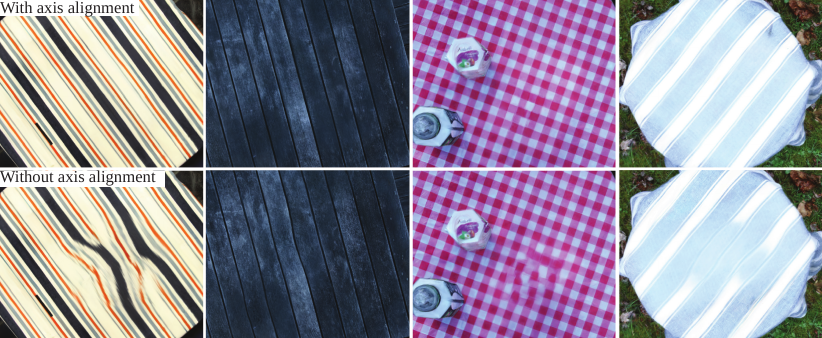}
    \caption{Qualitative visualization of the effect of our axis-alignment heuristic on inpainting results. Without it, the generation of context-consistent intricate patterns, being biased by the flow velocity towards alignment, is degraded.}
    \label{fig:axis-alignment-ablation-rasterized}
\end{figure}

As mentioned in \Cref{subsec:encoding_and_structure_completion}, we notice that TRELLIS' latent generation model $\mathcal G_L$ is more likely to generate lines and boundaries that are aligned with the axes of the voxel grid $O$ that associate a spatial position to the latent features. In this section, we first analyze that the axis-aligned bias comes from the latent flow model $\mathcal G_L$ rather than other stages of the pipeline. We then describe how we address this issue in practice by fitting the orientation of the voxel grid to visual features of the context $G_{c^*}$.

\paragraph{Axis aligned bias.} \Cref{fig:axis_bias} highlights and diagnoses the axis alignment bias of TRELLIS. First, we notice that the inpainting quality depends on the orientation of the \ac{SLAT} grid, which we further illustrate in \Cref{fig:axis-alignment-ablation-rasterized}. We quantify this degradation using LPIPS and show that this loss of quality does not depend on our attention masking \textbf{(a)}. We first rule out the possibility that the VAE (including structure prediction) causes the issue by empirically verifying that its auto-encoding is stable under rotation: we encode and decode the same scene under varying rotations around the vertical axis and measure no significant reconstruction error variation. However, we see in \textbf{(b)} that the latent flow model $\mathcal G_L$ is less reliable at predicting the correct flow velocity when the \ac{SLAT} grid is misaligned. Future versions of TRELLIS or a prior more robust to orientation changes could mitigate this issue and further improve our results.

\paragraph{Automatic orientation.} In practice, we assume that the vertical axis is already correct -- a wrong vertical axis causes many other issues, e.g., in terms of user navigation, and the vertical axis is a global reference, so it is usually correct and the assumption holds. 
The orientation angle around this vertical axis to align with the TRELLIS prior, however, is specific to the local neighborhood, namely the cropped context $G_{c^*}$. 
We estimate it by analyzing a $512\!\times\!512$ top-down view of this context. We first harmonize the rendered image by subtracting its low frequencies (by Gaussian-blurring the image, $\sigma=41$) and estimate the per-pixel dominant direction via the largest eigenvector of the local structure tensor (i.e., second-moment matrix) using a Sobel filter to estimate gradients. To improve the robustness of our estimation, we split the image into $8\!\times\!8$ tiles for which we compute the mean direction, then consider the median tile when sorting them by directional angle. To prevent the mask borders from impacting the orientation estimation, we zero the gradient in all pixels that lie within six pixels of the masked region.
We finally rotate voxel grid $O$ by this found angle around the vertical axis (or, equivalently, rotate the entire scene by the opposite angle).

\FloatBarrier

\section{Dataset \& model training}

\subsection{Sparse structure dataset curation}
\label{appendix:subcsec:structure-dataset}

 We curate an internal dataset of 900 3D Gaussian splatting scenes depicting full 360° environments rather than asset-scale captures. None of the evaluation scenes from COR-NeRF, Inpaint360, or 360-USID are included in this dataset. We select 100 candidate cropped context regions per scene. We sample a context region bounding box with a per-axis extent sampled randomly between 2\% and 80\% of the original scene's scale. Inside this context region, we further sample a target region bounding box, whose per-axis extent is randomly sampled between 10\% and 80\% of the context region. Additionally, we reject context/target region candidate pairs containing less than 20 000 and 5 000 Gaussian primitives respectively.  After sampling subscenes and accepting them based on the two criteria defined below, we apply data augmentation by horizontally flipping or rotating the structure by 90° increments, all with probability $0.25$. 
 
\paragraph{Objectness.}

The first criterion used to sample example pairs from the structure dataset depends on the presence or absence of objects in the target region, represented by local structure anomalies inside the masked region. We estimate the $3 \times 3 \times 3$ local covariance of every masked region voxel $\mathbf{p}$ and approximate the surface variation in the following way: \citep{pauly_efficient_2002}
$$\sigma(\mathbf{p}) = \frac{\lambda_3}{\lambda_1 + \lambda_2 + \lambda_3}$$
with $\lambda_1 \geq \lambda_2 \geq \lambda_3$ the eigenvalues of the local covariance matrix. Small values of $\sigma$ indicate locally planar structure.
We reject a subscene when more than 30\% of the target voxels show a variation $\sigma > 0.2$.

\paragraph{Structure predictability.}

In addition to the local structure covariance metric, we filter training dataset admissibility by computing target region predictability with respect to a quadric extrapolation of context. We partition the context structure $C$ into connected components and estimate the PCA local tangent frame of the component adjacent to the masked region \citep{hoppe_surface_1992}. Then, we fit a second-order polynomial height field \citep{cazals_estimating_2005} using context voxels:
$$f(u, v) = a_0 + a_1u + a_2v + a_3u^2 + a_4uv + a_5v^2$$

\begin{wrapfigure}{r}{0.5\linewidth}
    \vspace{-\intextsep}
    \centering
    \includegraphics[width=\linewidth]{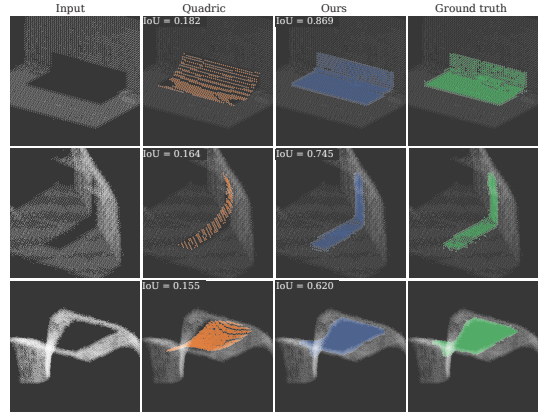}
    \caption{Quadric extrapolation against our learned structure prediction network for held-out test examples. Our model is able to successfully predict occupancy from complex context structure.}
    \label{fig:quadric-dataset}
\end{wrapfigure}

Finally, a target region is deemed admissible if the fraction of target voxels lying close to the extrapolated height field is higher than 0.5. We note that this deliberately permissive admissibility test simply serves as a way to filter target regions unexplainable by context out of the dataset, rather than selecting only target regions predictable by quadric extrapolation of context. The curated dataset contains a large number of non-flat target regions, such as corners or curved surfaces: \Cref{fig:quadric-dataset} displays representative examples from the held-out test set, accepted by our data curation process and complex to predict with quadric extrapolation. While classic techniques such as plane fitting with RANSAC~\citep{fischler_random_1981} could also be used to approximate individual examples, our learned structure prediction network is able to model occupancy completion from arbitrary context, without requiring any assumptions about its underlying structure.

\begin{figure}
    \centering
    \includegraphics{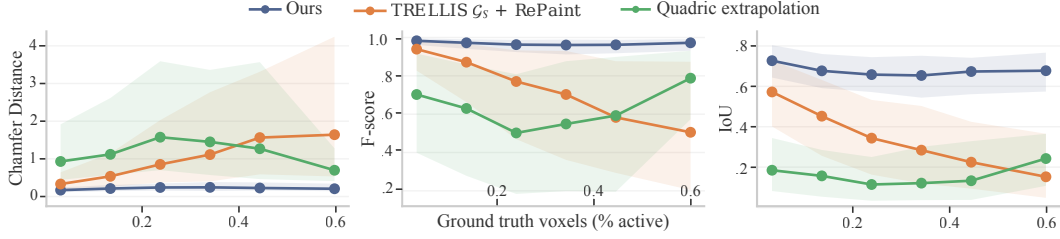}
    \caption{We ablate our proposed structure completion network with the TRELLIS backbone (see main text) and additionally show a comparison against a quadric-extrapolation baseline of the active ground truth voxels, which performs worse in all cases.}
    \label{fig:quadric_extrapolation}
\end{figure}

To verify this, we conduct an ablation study by comparing the respective metrics of our structure completion network and the quadric extrapolation baseline, on both held-out test 3D Gaussian splatting scenes and on the inpainting benchmark datasets. On the structure reconstruction task, our network outperforms quadric extrapolation over Chamfer distance, F-score and IoU metrics, indicating stronger sparse structure prediction capabilities (\Cref{fig:quadric_extrapolation}).

\needspace{0.5\textheight}

\subsection{Model training}
We divide the hereby constructed dataset into training, validation and test splits, using 650 scenes for training, 100 scenes for validation and 150 test scenes. Importantly, the train/test split is done at scene level and not at context/target region level, thus preventing data leakage. We train our model on 8 NVIDIA A100 GPUs for 300 epochs. We train using the Adam optimizer~\citep{kingma_adam_2015}, a batch size $B=16$, with a cosine-annealed learning rate between $2\times 10^{-4}$ and $2 \times 10^{-6}$. As mentioned in \Cref{subsec:encoding_and_structure_completion}, the loss function used for training is a linear combination of weighted cross-entropy and Dice loss:
$
    \mathcal{L}_\text{structure} = 0.3 \, \mathcal{L}_\text{W-BCE} + 0.7\, \mathcal{L}_\text{DICE}.
$
We use a weighted binary cross-entropy positive class weight of 100 during training. We conduct training with L2 weight regularization parameterized by $\lambda = 1 \times 10^{-4}$. After training for the full 300 epochs, the neural network weights achieving the lowest validation set loss are retained.

\section{Detailed per-scene results}

\Cref{tab:metric-human-agreement} reports human preference rating for each scene used in the user study. We then report a per-scene breakdown of perceptual metrics: Tables~\ref{tab:image-quality-cor}, \ref{tab:image-quality-inpaint360} and \ref{tab:image-quality-360-usid} detail the metrics of the four benchmarked methods, respectively on the COR-NeRF, Inpaint360 and 360-USID datasets.

\begin{figure}[h]
    \centering
    \includegraphics{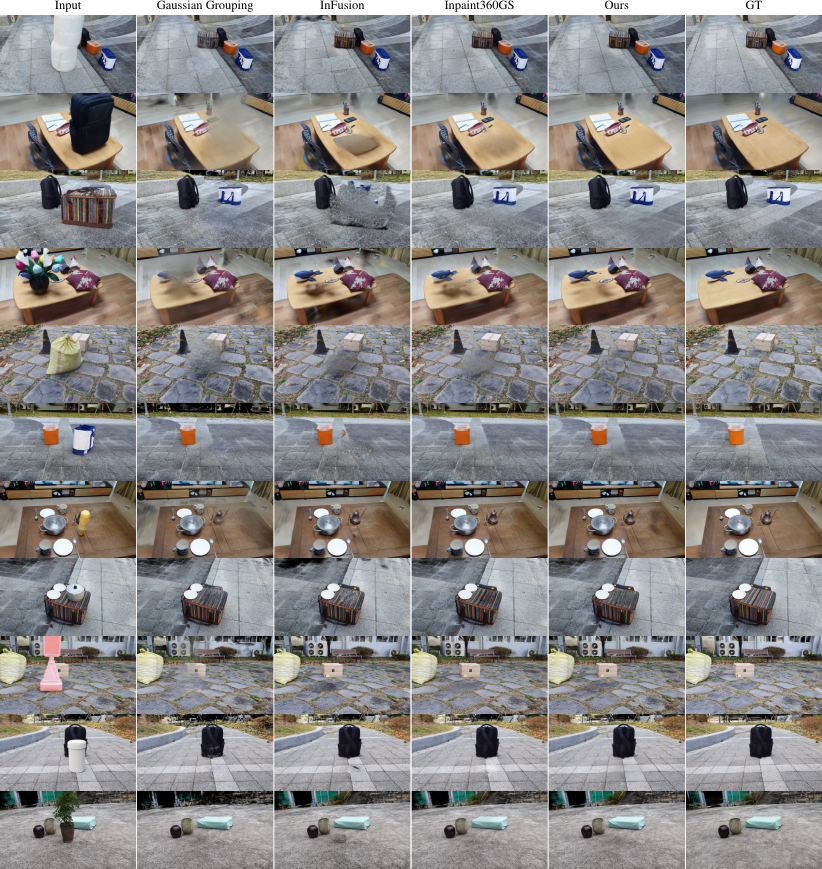}
    \caption{Qualitative comparison of our method and the implemented baselines on the COR-NeRF dataset. Our \textsc{Wings} framework provides sharp, consistent inpainting across the majority of the scenes in this dataset.}
    \label{fig:comparison-cor}
\end{figure}

\begin{figure}[t]
    \centering
    \includegraphics{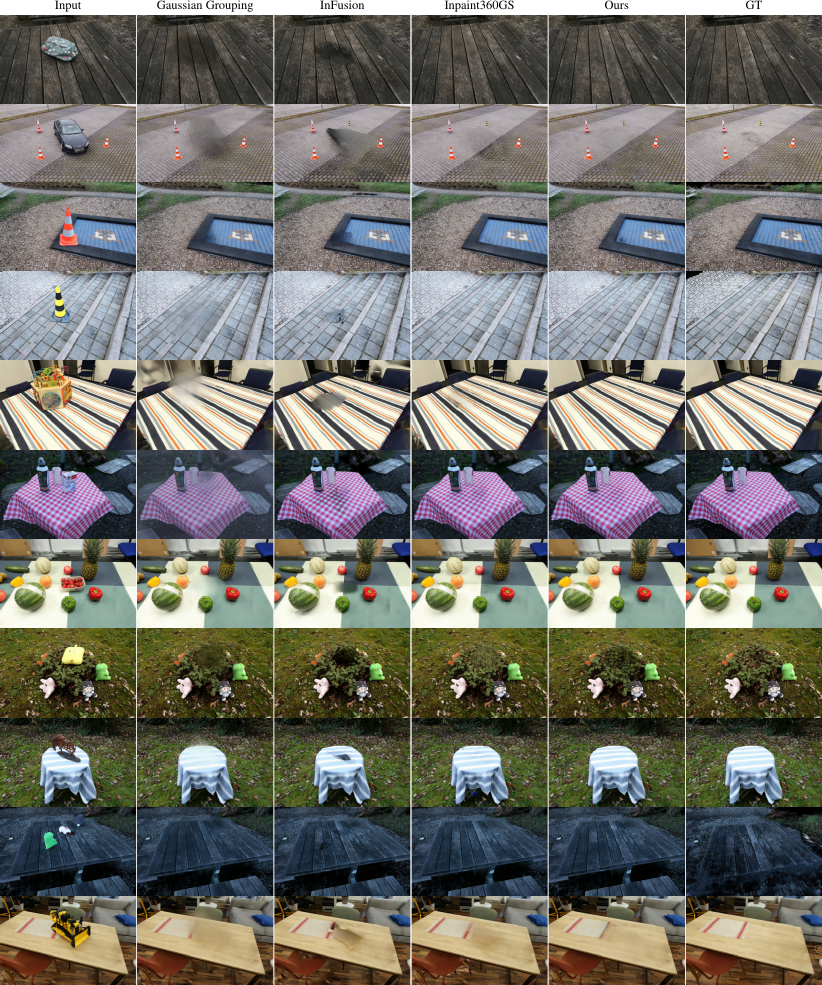}
    \caption{Qualitative comparison of our method and the implemented baselines on the Inpaint360 dataset.}
    \label{fig:comparison-inpaint360}
\end{figure}

\begin{figure}[t]
    \centering
    \includegraphics{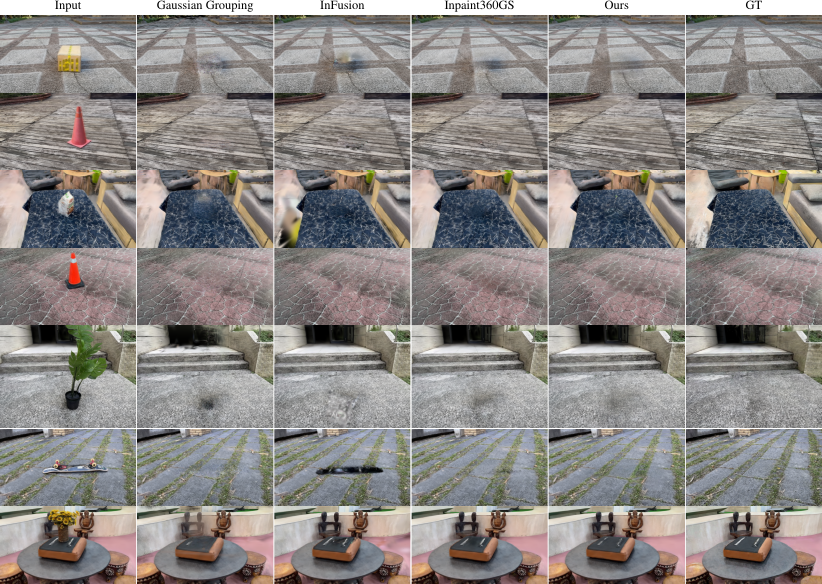}
    \caption{Qualitative comparison of our method and the implemented baselines on the 360-USID dataset.}
    \label{fig:comparison-usid}
\end{figure}

Finally, Figures~\ref{fig:comparison-cor}, \ref{fig:comparison-inpaint360} and \ref{fig:comparison-usid} show qualitative comparison grids of the four benchmarked methods over all scenes in the three evaluated datasets.% In addition, the reader is encouraged to visualize videos and additional comparison views in the electronic supplemental. %(removed for preprint)

\include{tables/human_metric_heatmap.tex}

%\FloatBarrier

\include{tables/COR_detailed_table}
\include{tables/inpaint360_detailed_table}
\include{tables/360-USID_detailed_table}

\FloatBarrier

\section{Detailed per-scene visualizations}
\label{appendix:subsec:per-scene-visualization}

\end{document}

%% file: math_commands.tex
\usepackage{amsmath,amsfonts,bm}

\def\eqref#1{equation~\ref{#1}}
\def\1{\bm{1}}

\DeclareMathAlphabet{\mathsfit}{\encodingdefault}{\sfdefault}{m}{sl}
\SetMathAlphabet{\mathsfit}{bold}{\encodingdefault}{\sfdefault}{bx}{n}

%% file: tables/human_metric_heatmap.tex
\definecolor{negstrong}{RGB}{204,46,74}
\definecolor{negmed}{RGB}{209,104,122}
\definecolor{neglight}{RGB}{214,162,171}
\definecolor{neutral}{RGB}{219,219,219}
\definecolor{poslight}{RGB}{174,203,181}
\definecolor{posmed}{RGB}{129,187,143}
\definecolor{posstrong}{RGB}{84,171,105}

\newcommand{\NS}{\cellcolor{negstrong}\strut}
\newcommand{\NM}{\cellcolor{negmed}\strut}
\newcommand{\NLT}{\cellcolor{neglight}\strut}
\newcommand{\ZZ}{\cellcolor{neutral}\strut}
\newcommand{\PL}{\cellcolor{poslight}\strut}
\newcommand{\PM}{\cellcolor{posmed}\strut}
\newcommand{\PS}{\cellcolor{posstrong}\strut}

\begin{table}[t]
    \centering
    \small
    \caption{
        Scene-level comparison of Ours and Inpaint360GS according to automated metrics and human preferences. Metric cells display within-metric normalized differences, where positive values take a green colour and favour Ours. The Human column reports the tie-adjusted preference for Ours. Across the 29 scenes, humans prefer Ours on 21, Inpaint360GS on 7 and rate the two as equal on one.
    }
    \label{tab:metric-human-agreement}
    \vspace{-2mm}

    \setlength{\tabcolsep}{2.5pt}
    \renewcommand{\arraystretch}{1.15}

    \resizebox{\columnwidth}{!}{%
    \begin{tabular}{llc@{\hspace{10pt}}ccccccc}
        \toprule
        & & & \multicolumn{3}{c}{Single-reference} & \multicolumn{4}{c}{Distributional} \\
        \cmidrule(lr){4-6}
        \cmidrule(lr){7-10}

        Dataset & Scene & \textbf{Human (\%)} & PSNR & SSIM & LPIPS & FID & Local-FID & KID & Local-KID \\
        \midrule

        \multirow{11}{*}{COR-NeRF}
        & aircap
        & \cellcolor{posmed}75.0
        & \NS & \PS & \PS & \PM & \PM & \ZZ & \PL \\

        & backpack
        & \cellcolor{posstrong}100.0
        & \PS & \PS & \PS & \PM & \PS & \NM & \PS \\

        & campingbox
        & \cellcolor{posstrong}88.9
        & \PS & \PM & \PS & \ZZ & \ZZ & \NM & \NLT \\

        & dolls
        & \cellcolor{poslight}66.7
        & \PS & \PS & \PS & \PS & \PS & \PS & \PS \\

        & gunnysack
        & \cellcolor{posstrong}100.0
        & \NS & \NS & \PM & \PS & \PL & \PS & \PL \\

        & icebox
        & \cellcolor{neutral}55.6
        & \NM & \NLT & \NLT & \ZZ & \NLT & \PL & \ZZ \\

        & mustard
        & \cellcolor{neglight}30.0
        & \NS & \NS & \NS & \NS & \NS & \NS & \NS \\

        & pot
        & \cellcolor{negmed}25.0
        & \PL & \NS & \NS & \NS & \NS & \NS & \NS \\

        & sign
        & \cellcolor{poslight}58.3
        & \NS & \ZZ & \PL & \ZZ & \PL & \PL & \PL \\

        & trashcan
        & \cellcolor{neglight}30.0
        & \PL & \ZZ & \PM & \ZZ & \PL & \ZZ & \ZZ \\

        & tree
        & \cellcolor{neglight}28.6
        & \ZZ & \NM & \PL & \PS & \PS & \PM & \PM \\

        \midrule

        \multirow{11}{*}{Inpaint360}
        & bag
        & \cellcolor{posmed}80.0
        & \NS & \NS & \NS & \NM & \NM & \NM & \ZZ \\

        & car
        & \cellcolor{posstrong}100.0
        & \NS & \NS & \PS & \PL & \PS & \NS & \NM \\

        & cone\_red
        & \cellcolor{posmed}75.0
        & \ZZ & \NM & \NM & \PS & \NS & \PS & \NM \\

        & cone\_yellow
        & \cellcolor{neutral}45.5
        & \PM & \NS & \NM & \NLT & \PS & \NLT & \PS \\

        & cube
        & \cellcolor{posstrong}100.0
        & \PS & \NLT & \PS & \PS & \PS & \PS & \PS \\

        & doppelherz
        & \cellcolor{posstrong}94.4
        & \PS & \PM & \NS & \PS & \PM & \PM & \PL \\

        & fruits
        & \cellcolor{negmed}21.4
        & \NS & \NM & \NS & \NS & \NS & \NM & \ZZ \\

        & garden\_toys
        & \cellcolor{posstrong}92.3
        & \NS & \NS & \NS & \NS & \NS & \NS & \NS \\

        & redbull
        & \cellcolor{posstrong}91.7
        & \PL & \ZZ & \PS & \ZZ & \PS & \NM & \PS \\

        & toys
        & \cellcolor{posmed}80.0
        & \PS & \PS & \ZZ & \NLT & \ZZ & \NM & \NS \\

        & truck
        & \cellcolor{neutral}50.0
        & \PS & \NM & \NS & \NS & \PS & \NS & \NM \\

        \midrule

        \multirow{7}{*}{360-USID}
        & carton
        & \cellcolor{posstrong}87.5
        & \PL & \NS & \PS & \PS & \PS & \PS & \PS \\

        & cone
        & \cellcolor{posstrong}96.2
        & \ZZ & \PS & \PS & \PS & \PS & \PS & \PS \\

        & cookie
        & \cellcolor{posmed}77.8
        & \NM & \NS & \NM & \PL & \NLT & \PS & \NS \\

        & newcone
        & \cellcolor{posmed}83.3
        & \PL & \NLT & \NS & \NS & \NM & \NS & \NS \\

        & plant
        & \cellcolor{neglight}37.5
        & \PM & \NS & \NM & \PS & \PM & \PS & \PS \\

        & skateboard
        & \cellcolor{posstrong}88.9
        & \PM & \PS & \ZZ & \PS & \PL & \PM & \PM \\

        & sunflower
        & \cellcolor{posstrong}90.0
        & \NS & \NM & \PL & \PS & \PS & \PS & \PS \\

        \midrule

        \multicolumn{3}{r}{Spearman $\rho$}
        & $+0.14$
        & $+0.21$
        & $+0.48$
        & $+0.40$
        & $+0.40$
        & $+0.13$
        & $+0.28$ \\

        \multicolumn{3}{r}{Sign agreement}
        & 54\%
        & 54\%
        & 64\%
        & 68\%
        & 61\%
        & 64\%
        & 54\% \\

        \bottomrule
    \end{tabular}%
    }

    \vspace{2mm}

    % Colour legend
    \begin{minipage}{0.84\columnwidth}
        \centering
        \footnotesize
        \begin{tabular}{@{}*{7}{p{0.12\linewidth}}@{}}
            \cellcolor{negstrong}\rule{0pt}{1.5ex} &
            \cellcolor{negmed} &
            \cellcolor{neglight} &
            \cellcolor{neutral} &
            \cellcolor{poslight} &
            \cellcolor{posmed} &
            \cellcolor{posstrong} \\
        \end{tabular}

        \vspace{0.5mm}

        \makebox[\linewidth][s]{%
            Inpaint360GS better
            \hfill \hspace{-10mm}
            parity
            \hfill
            Ours better
        }
    \end{minipage}

\end{table}

%% file: tables/COR_detailed_table.tex
\begin{table*}[t]
    \caption{
        Per-scene metrics on COR-NeRF. KID is reported
        $\times 10^{3}$. Human Pref. is each method's expected
        score against an average-skill opponent under the Davidson model, reported with a participant-bootstrap
        95\% confidence interval. Bold denotes the best method for each scene and metric.
    }
    \label{tab:image-quality-cor}
    \centering
    \small
    \setlength{\tabcolsep}{3.5pt}
    \renewcommand{\arraystretch}{1.04}
    \resizebox{\linewidth}{!}{%
    \begin{tabular}{llccc@{\hspace{5pt}}ccc@{\hspace{5pt}}c}
        \toprule
         & & \multicolumn{3}{c}{\textbf{Reference-based}} & \multicolumn{3}{c}{\textbf{Distributional}} & \multicolumn{1}{c}{\textbf{Human}} \\
        \cmidrule(lr){3-5} \cmidrule(lr){6-8} \cmidrule(l){9-9}
        \textbf{Scene} & \textbf{Method} & \textbf{PSNR} $\uparrow$ & \textbf{SSIM} $\uparrow$ & \textbf{LPIPS} $\downarrow$ & \textbf{FID} $\downarrow$ & \textbf{Local-FID} $\downarrow$ & \textbf{KID} $\downarrow$ & \textbf{Human Pref.} $\uparrow$ \\
        \midrule
\multirow{4}{*}{\textbf{Aircap}} & Gaussian Grouping & $20.35$ & $0.599$ & $0.259$ & $44.04$ & $41.31$ & $4.26$ & $48.3 \pm 10.7$ \\
 & InFusion & $19.99$ & $0.598$ & $0.260$ & $42.70$ & $40.44$ & $5.17$ & $23.0 \pm 10.7$ \\
 & Inpaint360GS & $\bm{21.41}$ & $0.612$ & $0.251$ & $37.65$ & $34.47$ & $2.28$ & $63.4 \pm 11.2$ \\
 & Ours & $21.13$ & $\bm{0.617}$ & $\bm{0.247}$ & $\bm{36.55}$ & $\bm{32.36}$ & $\bm{2.12}$ & $\bm{67.8 \pm 13.9}$ \\
\addlinespace[3pt]
\multirow{4}{*}{\textbf{Backpack}} & Gaussian Grouping & $21.34$ & $0.856$ & $0.211$ & $60.66$ & $57.02$ & $13.40$ & $16.2 \pm 5.4$ \\
 & InFusion & $21.59$ & $0.877$ & $0.184$ & $66.87$ & $66.63$ & $21.88$ & $17.8 \pm 8.7$ \\
 & Inpaint360GS & $23.04$ & $0.899$ & $0.159$ & $40.91$ & $42.19$ & $\bm{4.57}$ & $65.0 \pm 8.8$ \\
 & Ours & $\bm{23.51}$ & $\bm{0.905}$ & $\bm{0.152}$ & $\bm{39.34}$ & $\bm{34.95}$ & $5.49$ & $\bm{92.4 \pm 1.5}$ \\
\addlinespace[3pt]
\multirow{4}{*}{\textbf{Campingbox}} & Gaussian Grouping & $21.72$ & $0.583$ & $0.260$ & $39.59$ & $32.22$ & $6.47$ & $49.7 \pm 8.6$ \\
 & InFusion & $20.21$ & $0.575$ & $0.287$ & $124.72$ & $121.77$ & $79.75$ & $10.7 \pm 3.7$ \\
 & Inpaint360GS & $22.95$ & $0.608$ & $0.247$ & $\bm{29.42}$ & $26.23$ & $\bm{1.36}$ & $64.6 \pm 6.9$ \\
 & Ours & $\bm{23.14}$ & $\bm{0.610}$ & $\bm{0.244}$ & $29.62$ & $\bm{26.16}$ & $2.72$ & $\bm{82.9 \pm 7.3}$ \\
\addlinespace[3pt]
\multirow{4}{*}{\textbf{Dolls}} & Gaussian Grouping & $20.79$ & $0.844$ & $0.204$ & $70.06$ & $63.51$ & $15.34$ & $25.5 \pm 8.2$ \\
 & InFusion & $21.14$ & $0.865$ & $0.173$ & $60.73$ & $52.64$ & $9.95$ & $42.9 \pm 6.1$ \\
 & Inpaint360GS & $23.32$ & $0.889$ & $0.159$ & $57.95$ & $62.16$ & $7.60$ & $53.8 \pm 6.3$ \\
 & Ours & $\bm{25.27}$ & $\bm{0.904}$ & $\bm{0.139}$ & $\bm{43.31}$ & $\bm{37.64}$ & $\bm{3.61}$ & $\bm{77.0 \pm 8.0}$ \\
\addlinespace[3pt]
\multirow{4}{*}{\textbf{Gunnysack}} & Gaussian Grouping & $19.71$ & $0.628$ & $0.236$ & $45.65$ & $43.26$ & $7.41$ & $24.0 \pm 9.4$ \\
 & InFusion & $19.21$ & $0.648$ & $0.218$ & $34.38$ & $33.67$ & $3.12$ & $30.8 \pm 10.1$ \\
 & Inpaint360GS & $\bm{20.35}$ & $\bm{0.654}$ & $0.216$ & $34.41$ & $32.52$ & $3.21$ & $43.3 \pm 8.1$ \\
 & Ours & $20.17$ & $0.650$ & $\bm{0.214}$ & $\bm{32.01}$ & $\bm{31.44}$ & $\bm{1.46}$ & $\bm{89.7 \pm 2.4}$ \\
\addlinespace[3pt]
\multirow{4}{*}{\textbf{Icebox}} & Gaussian Grouping & $21.15$ & $0.529$ & $0.297$ & $31.46$ & $26.59$ & $3.53$ & $45.9 \pm 14.6$ \\
 & InFusion & $22.21$ & $0.547$ & $0.293$ & $34.50$ & $34.44$ & $5.15$ & $19.3 \pm 6.4$ \\
 & Inpaint360GS & $\bm{23.15}$ & $\bm{0.558}$ & $\bm{0.281}$ & $26.07$ & $\bm{22.55}$ & $1.97$ & $65.7 \pm 14.0$ \\
 & Ours & $23.01$ & $0.556$ & $\bm{0.281}$ & $\bm{25.93}$ & $22.88$ & $\bm{1.59}$ & $\bm{72.4 \pm 10.7}$ \\
\addlinespace[3pt]
\multirow{4}{*}{\textbf{Mustard}} & Gaussian Grouping & $18.61$ & $0.786$ & $0.238$ & $68.07$ & $46.90$ & $17.59$ & $36.1 \pm 9.4$ \\
 & InFusion & $21.65$ & $0.839$ & $0.196$ & $42.89$ & $33.72$ & $4.58$ & $40.0 \pm 9.8$ \\
 & Inpaint360GS & $\bm{23.37}$ & $\bm{0.863}$ & $\bm{0.175}$ & $\bm{36.12}$ & $\bm{30.15}$ & $\bm{0.99}$ & $\bm{67.2 \pm 8.7}$ \\
 & Ours & $21.31$ & $0.833$ & $0.196$ & $44.15$ & $33.77$ & $6.77$ & $56.4 \pm 8.7$ \\
\addlinespace[3pt]
\multirow{4}{*}{\textbf{Pot}} & Gaussian Grouping & $19.65$ & $0.575$ & $0.255$ & $65.31$ & $65.84$ & $15.75$ & $44.5 \pm 12.9$ \\
 & InFusion & $19.49$ & $0.578$ & $0.254$ & $58.79$ & $56.79$ & $12.73$ & $34.1 \pm 14.1$ \\
 & Inpaint360GS & $20.29$ & $\bm{0.593}$ & $\bm{0.242}$ & $\bm{47.63}$ & $\bm{46.25}$ & $\bm{7.20}$ & $\bm{66.7 \pm 14.1}$ \\
 & Ours & $\bm{20.34}$ & $0.588$ & $0.246$ & $60.88$ & $64.06$ & $15.39$ & $54.6 \pm 12.8$ \\
\addlinespace[3pt]
\multirow{4}{*}{\textbf{Sign}} & Gaussian Grouping & $21.22$ & $0.691$ & $0.188$ & $33.60$ & $30.35$ & $2.85$ & $30.1 \pm 13.5$ \\
 & InFusion & $21.37$ & $0.704$ & $0.177$ & $27.26$ & $24.32$ & $1.18$ & $43.7 \pm 12.9$ \\
 & Inpaint360GS & $\bm{22.18}$ & $\bm{0.712}$ & $0.173$ & $26.22$ & $23.48$ & $0.79$ & $60.9 \pm 12.6$ \\
 & Ours & $21.93$ & $0.711$ & $\bm{0.172}$ & $\bm{26.16}$ & $\bm{23.16}$ & $\bm{0.41}$ & $\bm{65.8 \pm 11.7}$ \\
\addlinespace[3pt]
\multirow{4}{*}{\textbf{Trashcan}} & Gaussian Grouping & $21.96$ & $0.585$ & $0.217$ & $28.83$ & $21.79$ & $0.95$ & $52.6 \pm 12.2$ \\
 & InFusion & $22.65$ & $0.598$ & $0.213$ & $28.28$ & $20.37$ & $1.40$ & $26.4 \pm 11.8$ \\
 & Inpaint360GS & $23.28$ & $\bm{0.604}$ & $0.207$ & $\bm{23.95}$ & $18.40$ & $\bm{-0.39}$ & $\bm{62.1 \pm 13.6}$ \\
 & Ours & $\bm{23.33}$ & $\bm{0.604}$ & $\bm{0.205}$ & $\bm{23.95}$ & $\bm{18.13}$ & $-0.22$ & $60.7 \pm 13.7$ \\
\addlinespace[3pt]
\multirow{4}{*}{\textbf{Tree}} & Gaussian Grouping & $21.38$ & $0.582$ & $0.282$ & $41.35$ & $34.79$ & $2.53$ & $\bm{71.0 \pm 10.3}$ \\
 & InFusion & $22.39$ & $0.614$ & $0.264$ & $37.31$ & $33.12$ & $2.20$ & $22.0 \pm 9.5$ \\
 & Inpaint360GS & $\bm{23.27}$ & $\bm{0.627}$ & $\bm{0.256}$ & $36.71$ & $32.63$ & $1.41$ & $63.4 \pm 11.0$ \\
 & Ours & $\bm{23.27}$ & $0.625$ & $\bm{0.256}$ & $\bm{34.26}$ & $\bm{29.46}$ & $\bm{0.53}$ & $45.8 \pm 7.8$ \\
\midrule
\multirow{4}{*}{\textbf{Average}} & Gaussian Grouping & $20.72$ & $0.660$ & $0.241$ & $48.06$ & $42.14$ & $8.19$ & $39.6 \pm 4.6$ \\
 & InFusion & $21.08$ & $0.677$ & $0.229$ & $50.77$ & $47.08$ & $13.37$ & $27.5 \pm 3.3$ \\
 & Inpaint360GS & $\bm{22.42}$ & $\bm{0.693}$ & $0.215$ & $36.10$ & $33.73$ & $\bm{2.82}$ & $61.2 \pm 3.4$ \\
 & Ours & $22.40$ & $0.691$ & $\bm{0.214}$ & $\bm{36.02}$ & $\bm{32.18}$ & $3.63$ & $\bm{71.8 \pm 4.1}$ \\
\addlinespace[3pt]
\multirow{4}{*}{\textbf{Median}} & Gaussian Grouping & $21.15$ & $0.599$ & $0.238$ & $44.04$ & $41.31$ & $6.47$ & \textemdash \\
 & InFusion & $21.37$ & $0.614$ & $0.218$ & $42.70$ & $34.44$ & $5.15$ & \textemdash \\
 & Inpaint360GS & $\bm{23.04}$ & $\bm{0.627}$ & $0.216$ & $36.12$ & $32.52$ & $\bm{1.97}$ & \textemdash \\
 & Ours & $23.01$ & $0.625$ & $\bm{0.214}$ & $\bm{34.26}$ & $\bm{31.44}$ & $2.12$ & \textemdash \\
        \bottomrule
    \end{tabular}
    }
\end{table*}

%% file: tables/inpaint360_detailed_table.tex
\begin{table*}[t]
    \caption{
        Per-scene metrics on Inpaint360. KID is reported
        $\times 10^{3}$. Human Pref. is each method's expected
        score against an average-skill opponent under the Davidson model, reported with a participant-bootstrap
        95\% confidence interval. Bold denotes the best method for each scene and metric.
    }
    \label{tab:image-quality-inpaint360}
    \centering
    \small
    \setlength{\tabcolsep}{3.5pt}
    \renewcommand{\arraystretch}{1.04}
    \resizebox{\linewidth}{!}{%
    \begin{tabular}{llccc@{\hspace{5pt}}ccc@{\hspace{5pt}}c}
        \toprule
         & & \multicolumn{3}{c}{\textbf{Reference-based}} & \multicolumn{3}{c}{\textbf{Distributional}} & \multicolumn{1}{c}{\textbf{Human}} \\
        \cmidrule(lr){3-5} \cmidrule(lr){6-8} \cmidrule(l){9-9}
        \textbf{Scene} & \textbf{Method} & \textbf{PSNR} $\uparrow$ & \textbf{SSIM} $\uparrow$ & \textbf{LPIPS} $\downarrow$ & \textbf{FID} $\downarrow$ & \textbf{Local-FID} $\downarrow$ & \textbf{KID} $\downarrow$ & \textbf{Human Pref.} $\uparrow$ \\
        \midrule
\multirow{4}{*}{\textbf{Bag}} & Gaussian Grouping & $24.84$ & $0.835$ & $0.148$ & $35.46$ & $35.38$ & $-0.55$ & $25.9 \pm 6.9$ \\
 & InFusion & $25.16$ & $0.842$ & $0.157$ & $60.34$ & $70.92$ & $15.04$ & $10.7 \pm 4.3$ \\
 & Inpaint360GS & $\bm{27.47}$ & $\bm{0.856}$ & $\bm{0.133}$ & $\bm{29.08}$ & $\bm{25.85}$ & $\bm{-3.55}$ & $75.0 \pm 7.8$ \\
 & Ours & $27.20$ & $0.848$ & $0.136$ & $30.28$ & $27.20$ & $-2.44$ & $\bm{88.8 \pm 5.1}$ \\
\addlinespace[3pt]
\multirow{4}{*}{\textbf{Car}} & Gaussian Grouping & $18.91$ & $0.759$ & $0.184$ & $127.19$ & $122.27$ & $66.05$ & $29.9 \pm 5.7$ \\
 & InFusion & $17.58$ & $0.721$ & $0.217$ & $171.26$ & $143.10$ & $108.98$ & $25.7 \pm 6.3$ \\
 & Inpaint360GS & $\bm{21.74}$ & $\bm{0.774}$ & $0.176$ & $81.52$ & $83.63$ & $\bm{16.21}$ & $53.6 \pm 6.7$ \\
 & Ours & $21.55$ & $0.740$ & $\bm{0.172}$ & $\bm{79.75}$ & $\bm{67.69}$ & $27.74$ & $\bm{85.7 \pm 5.5}$ \\
\addlinespace[3pt]
\multirow{4}{*}{\textbf{Cone Red}} & Gaussian Grouping & $20.46$ & $0.726$ & $0.189$ & $33.16$ & $30.22$ & $6.30$ & $28.1 \pm 6.9$ \\
 & InFusion & $20.29$ & $0.726$ & $0.197$ & $58.43$ & $53.38$ & $13.31$ & $11.0 \pm 5.2$ \\
 & Inpaint360GS & $20.60$ & $\bm{0.736}$ & $\bm{0.182}$ & $22.36$ & $\bm{18.42}$ & $0.32$ & $76.2 \pm 8.2$ \\
 & Ours & $\bm{20.61}$ & $0.733$ & $0.184$ & $\bm{21.23}$ & $20.81$ & $\bm{-1.44}$ & $\bm{86.7 \pm 7.2}$ \\
\addlinespace[3pt]
\multirow{4}{*}{\textbf{Cone Yellow}} & Gaussian Grouping & $21.23$ & $0.727$ & $0.192$ & $46.08$ & $44.67$ & $7.03$ & $32.4 \pm 6.7$ \\
 & InFusion & $21.44$ & $0.774$ & $0.171$ & $128.75$ & $130.81$ & $67.78$ & $10.1 \pm 2.8$ \\
 & Inpaint360GS & $21.78$ & $\bm{0.782}$ & $\bm{0.155}$ & $\bm{27.79}$ & $24.30$ & $\bm{-9.04}$ & $\bm{82.6 \pm 5.1}$ \\
 & Ours & $\bm{21.90}$ & $0.777$ & $0.157$ & $28.41$ & $\bm{19.42}$ & $-8.42$ & $80.3 \pm 5.7$ \\
\addlinespace[3pt]
\multirow{4}{*}{\textbf{Cube}} & Gaussian Grouping & $19.52$ & $0.802$ & $0.195$ & $90.24$ & $118.28$ & $31.50$ & $35.4 \pm 8.9$ \\
 & InFusion & $18.63$ & $0.831$ & $0.175$ & $88.03$ & $110.25$ & $23.85$ & $16.3 \pm 6.4$ \\
 & Inpaint360GS & $22.22$ & $\bm{0.866}$ & $0.143$ & $45.71$ & $43.89$ & $-1.15$ & $49.0 \pm 7.6$ \\
 & Ours & $\bm{22.55}$ & $0.864$ & $\bm{0.122}$ & $\bm{35.46}$ & $\bm{23.47}$ & $\bm{-4.87}$ & $\bm{90.2 \pm 3.4}$ \\
\addlinespace[3pt]
\multirow{4}{*}{\textbf{Doppelherz}} & Gaussian Grouping & $22.65$ & $0.790$ & $0.146$ & $53.23$ & $49.82$ & $-6.91$ & $17.5 \pm 6.6$ \\
 & InFusion & $24.31$ & $0.827$ & $0.126$ & $49.59$ & $45.11$ & $-4.29$ & $24.6 \pm 9.0$ \\
 & Inpaint360GS & $25.42$ & $0.843$ & $\bm{0.118}$ & $39.21$ & $29.92$ & $-10.29$ & $61.4 \pm 10.0$ \\
 & Ours & $\bm{26.06}$ & $\bm{0.846}$ & $0.119$ & $\bm{37.12}$ & $\bm{28.53}$ & $\bm{-11.47}$ & $\bm{89.7 \pm 4.2}$ \\
\addlinespace[3pt]
\multirow{4}{*}{\textbf{Fruits}} & Gaussian Grouping & $24.28$ & $0.866$ & $0.171$ & $35.62$ & $28.05$ & $-5.97$ & $26.1 \pm 9.0$ \\
 & InFusion & $23.89$ & $0.869$ & $0.166$ & $31.11$ & $25.44$ & $\bm{-8.47}$ & $23.4 \pm 8.9$ \\
 & Inpaint360GS & $\bm{25.42}$ & $\bm{0.879}$ & $\bm{0.158}$ & $\bm{29.05}$ & $\bm{23.14}$ & $-8.10$ & $\bm{85.1 \pm 6.2}$ \\
 & Ours & $25.16$ & $0.875$ & $0.162$ & $31.42$ & $25.32$ & $-6.88$ & $61.3 \pm 15.3$ \\
\addlinespace[3pt]
\multirow{4}{*}{\textbf{Garden Toys}} & Gaussian Grouping & $23.12$ & $0.752$ & $0.186$ & $53.02$ & $45.44$ & $-5.52$ & $43.6 \pm 9.7$ \\
 & InFusion & $22.52$ & $0.756$ & $0.185$ & $55.04$ & $44.48$ & $-2.02$ & $12.4 \pm 5.0$ \\
 & Inpaint360GS & $\bm{23.58}$ & $\bm{0.763}$ & $\bm{0.178}$ & $\bm{43.49}$ & $\bm{40.59}$ & $\bm{-10.09}$ & $51.2 \pm 13.1$ \\
 & Ours & $23.14$ & $0.747$ & $0.189$ & $55.23$ & $46.80$ & $-3.20$ & $\bm{89.5 \pm 4.3}$ \\
\addlinespace[3pt]
\multirow{4}{*}{\textbf{Redbull}} & Gaussian Grouping & $23.71$ & $0.751$ & $0.185$ & $46.48$ & $46.20$ & $14.73$ & $29.1 \pm 6.0$ \\
 & InFusion & $22.90$ & $0.748$ & $0.186$ & $89.76$ & $100.81$ & $50.38$ & $9.4 \pm 4.2$ \\
 & Inpaint360GS & $24.07$ & $\bm{0.756}$ & $0.176$ & $\bm{31.87}$ & $34.55$ & $\bm{2.47}$ & $70.4 \pm 9.3$ \\
 & Ours & $\bm{24.15}$ & $\bm{0.756}$ & $\bm{0.174}$ & $32.05$ & $\bm{29.20}$ & $3.46$ & $\bm{90.8 \pm 3.2}$ \\
\addlinespace[3pt]
\multirow{4}{*}{\textbf{Toys}} & Gaussian Grouping & $\bm{18.58}$ & $\bm{0.237}$ & $0.527$ & $100.35$ & $76.14$ & $58.16$ & $64.6 \pm 8.9$ \\
 & InFusion & $18.22$ & $0.208$ & $0.535$ & $103.88$ & $90.94$ & $53.91$ & $13.3 \pm 3.5$ \\
 & Inpaint360GS & $17.92$ & $0.216$ & $0.526$ & $\bm{93.21}$ & $\bm{65.85}$ & $\bm{52.96}$ & $53.1 \pm 10.6$ \\
 & Ours & $18.14$ & $0.225$ & $\bm{0.525}$ & $94.61$ & $66.16$ & $53.98$ & $\bm{76.6 \pm 7.6}$ \\
\addlinespace[3pt]
\multirow{4}{*}{\textbf{Truck}} & Gaussian Grouping & $20.87$ & $0.837$ & $0.180$ & $59.84$ & $70.15$ & $19.72$ & $30.2 \pm 6.8$ \\
 & InFusion & $22.35$ & $0.861$ & $0.162$ & $88.96$ & $105.32$ & $54.61$ & $29.9 \pm 7.5$ \\
 & Inpaint360GS & $24.11$ & $\bm{0.876}$ & $\bm{0.144}$ & $\bm{38.15}$ & $50.94$ & $\bm{4.37}$ & $64.0 \pm 9.8$ \\
 & Ours & $\bm{24.60}$ & $0.872$ & $0.149$ & $40.20$ & $\bm{46.71}$ & $7.24$ & $\bm{75.3 \pm 7.8}$ \\
\midrule
\multirow{4}{*}{\textbf{Average}} & Gaussian Grouping & $21.65$ & $0.735$ & $0.209$ & $61.88$ & $60.60$ & $16.78$ & $29.7 \pm 5.4$ \\
 & InFusion & $21.57$ & $0.742$ & $0.207$ & $84.10$ & $83.69$ & $33.92$ & $12.4 \pm 3.5$ \\
 & Inpaint360GS & $23.12$ & $\bm{0.759}$ & $\bm{0.190}$ & $\bm{43.77}$ & $40.10$ & $\bm{3.10}$ & $69.3 \pm 4.8$ \\
 & Ours & $\bm{23.19}$ & $0.753$ & $\bm{0.190}$ & $44.16$ & $\bm{36.48}$ & $4.88$ & $\bm{88.1 \pm 3.7}$ \\
\addlinespace[3pt]
\multirow{4}{*}{\textbf{Median}} & Gaussian Grouping & $21.23$ & $0.759$ & $0.185$ & $53.02$ & $46.20$ & $7.03$ & \textemdash \\
 & InFusion & $22.35$ & $0.774$ & $0.175$ & $88.03$ & $90.94$ & $23.85$ & \textemdash \\
 & Inpaint360GS & $\bm{23.58}$ & $\bm{0.782}$ & $\bm{0.158}$ & $38.15$ & $34.55$ & $-1.15$ & \textemdash \\
 & Ours & $23.14$ & $0.777$ & $0.162$ & $\bm{35.46}$ & $\bm{28.53}$ & $\bm{-2.44}$ & \textemdash \\
        \bottomrule
    \end{tabular}
    }
\end{table*}

%% file: tables/360-USID_detailed_table.tex
\begin{table*}[t]
    \caption{
        Per-scene metrics on 360-USID. KID is reported
        $\times 10^{3}$. Human Pref. is each method's expected
        score against an average-skill opponent under the Davidson model, reported with a participant-bootstrap
        95\% confidence interval. Bold denotes the best method for each scene and metric.
    }
    \label{tab:image-quality-360-usid}
    \centering
    \small
    \setlength{\tabcolsep}{3.5pt}
    \renewcommand{\arraystretch}{1.04}
    \resizebox{\linewidth}{!}{%
    \begin{tabular}{llccc@{\hspace{5pt}}ccc@{\hspace{5pt}}c}
        \toprule
         & & \multicolumn{3}{c}{\textbf{Reference-based}} & \multicolumn{3}{c}{\textbf{Distributional}} & \multicolumn{1}{c}{\textbf{Human}} \\
        \cmidrule(lr){3-5} \cmidrule(lr){6-8} \cmidrule(l){9-9}
        \textbf{Scene} & \textbf{Method} & \textbf{PSNR} $\uparrow$ & \textbf{SSIM} $\uparrow$ & \textbf{LPIPS} $\downarrow$ & \textbf{FID} $\downarrow$ & \textbf{Local-FID} $\downarrow$ & \textbf{KID} $\downarrow$ & \textbf{Human Pref.} $\uparrow$ \\
        \midrule
\multirow{4}{*}{\textbf{Carton}} & Gaussian Grouping & $20.46$ & $0.459$ & $0.324$ & $44.84$ & $68.15$ & $15.06$ & $36.1 \pm 10.5$ \\
 & InFusion & $21.01$ & $0.498$ & $0.309$ & $87.81$ & $128.00$ & $59.90$ & $12.1 \pm 5.2$ \\
 & Inpaint360GS & $21.38$ & $\bm{0.499}$ & $0.302$ & $37.63$ & $63.23$ & $11.52$ & $72.0 \pm 10.3$ \\
 & Ours & $\bm{21.44}$ & $0.495$ & $\bm{0.298}$ & $\bm{34.51}$ & $\bm{53.13}$ & $\bm{8.04}$ & $\bm{83.7 \pm 8.2}$ \\
\addlinespace[3pt]
\multirow{4}{*}{\textbf{Cone}} & Gaussian Grouping & $16.49$ & $0.183$ & $\bm{0.319}$ & $55.10$ & $67.60$ & $2.71$ & $\bm{77.4 \pm 10.0}$ \\
 & InFusion & $16.45$ & $0.184$ & $0.326$ & $68.07$ & $84.39$ & $7.95$ & $27.9 \pm 10.0$ \\
 & Inpaint360GS & $16.55$ & $0.186$ & $0.323$ & $61.02$ & $83.75$ & $4.22$ & $28.6 \pm 9.8$ \\
 & Ours & $\bm{16.56}$ & $\bm{0.187}$ & $\bm{0.319}$ & $\bm{51.26}$ & $\bm{64.72}$ & $\bm{-0.08}$ & $65.1 \pm 9.4$ \\
\addlinespace[3pt]
\multirow{4}{*}{\textbf{Cookie}} & Gaussian Grouping & $18.11$ & $0.478$ & $0.366$ & $195.05$ & $218.27$ & $79.21$ & $19.4 \pm 8.2$ \\
 & InFusion & $17.79$ & $0.480$ & $0.365$ & $195.85$ & $188.97$ & $61.92$ & $56.1 \pm 11.2$ \\
 & Inpaint360GS & $\bm{19.44}$ & $\bm{0.508}$ & $\bm{0.349}$ & $161.61$ & $\bm{179.79}$ & $43.58$ & $55.0 \pm 11.0$ \\
 & Ours & $19.32$ & $0.505$ & $0.351$ & $\bm{158.88}$ & $185.81$ & $\bm{33.72}$ & $\bm{73.1 \pm 10.3}$ \\
\addlinespace[3pt]
\multirow{4}{*}{\textbf{Newcone}} & Gaussian Grouping & $21.62$ & $0.447$ & $0.311$ & $88.21$ & $75.68$ & $27.26$ & $31.4 \pm 9.7$ \\
 & InFusion & $21.48$ & $0.447$ & $0.309$ & $96.69$ & $86.83$ & $44.25$ & $29.1 \pm 9.6$ \\
 & Inpaint360GS & $21.64$ & $\bm{0.459}$ & $\bm{0.301}$ & $\bm{77.33}$ & $\bm{69.06}$ & $\bm{14.77}$ & $60.7 \pm 10.7$ \\
 & Ours & $\bm{21.67}$ & $0.458$ & $0.307$ & $81.49$ & $74.78$ & $21.56$ & $\bm{77.3 \pm 10.3}$ \\
\addlinespace[3pt]
\multirow{4}{*}{\textbf{Plant}} & Gaussian Grouping & $19.02$ & $0.508$ & $0.319$ & $83.89$ & $77.67$ & $22.30$ & $29.5 \pm 10.9$ \\
 & InFusion & $18.65$ & $0.490$ & $0.341$ & $105.98$ & $131.25$ & $44.07$ & $16.4 \pm 7.3$ \\
 & Inpaint360GS & $19.12$ & $\bm{0.514}$ & $\bm{0.312}$ & $49.79$ & $56.35$ & $1.55$ & $77.3 \pm 11.0$ \\
 & Ours & $\bm{19.25}$ & $0.500$ & $0.315$ & $\bm{44.94}$ & $\bm{54.00}$ & $\bm{-3.22}$ & $\bm{78.3 \pm 8.3}$ \\
\addlinespace[3pt]
\multirow{4}{*}{\textbf{Skateboard}} & Gaussian Grouping & $19.87$ & $0.378$ & $\bm{0.300}$ & $38.11$ & $40.12$ & $3.07$ & $62.5 \pm 10.6$ \\
 & InFusion & $17.52$ & $0.378$ & $0.316$ & $256.83$ & $194.96$ & $271.66$ & $9.8 \pm 3.7$ \\
 & Inpaint360GS & $19.90$ & $0.383$ & $0.303$ & $38.45$ & $40.79$ & $3.07$ & $49.1 \pm 12.2$ \\
 & Ours & $\bm{20.02}$ & $\bm{0.387}$ & $0.303$ & $\bm{35.66}$ & $\bm{39.61}$ & $\bm{1.69}$ & $\bm{85.6 \pm 7.9}$ \\
\addlinespace[3pt]
\multirow{4}{*}{\textbf{Sunflower}} & Gaussian Grouping & $19.98$ & $0.745$ & $0.262$ & $131.87$ & $130.43$ & $54.05$ & $24.9 \pm 10.9$ \\
 & InFusion & $22.74$ & $0.784$ & $0.238$ & $107.69$ & $112.66$ & $25.68$ & $39.5 \pm 11.0$ \\
 & Inpaint360GS & $\bm{23.85}$ & $\bm{0.805}$ & $0.222$ & $89.64$ & $91.26$ & $9.11$ & $53.0 \pm 12.4$ \\
 & Ours & $23.63$ & $0.802$ & $\bm{0.221}$ & $\bm{81.22}$ & $\bm{80.68}$ & $\bm{6.45}$ & $\bm{80.2 \pm 10.5}$ \\
\midrule
\multirow{4}{*}{\textbf{Average}} & Gaussian Grouping & $19.36$ & $0.457$ & $0.314$ & $91.01$ & $96.85$ & $29.10$ & $37.0 \pm 4.5$ \\
 & InFusion & $19.38$ & $0.466$ & $0.315$ & $131.27$ & $132.44$ & $73.63$ & $27.0 \pm 5.2$ \\
 & Inpaint360GS & $\bm{20.27}$ & $\bm{0.479}$ & $\bm{0.302}$ & $73.64$ & $83.46$ & $12.55$ & $53.4 \pm 4.8$ \\
 & Ours & $\bm{20.27}$ & $0.476$ & $\bm{0.302}$ & $\bm{69.71}$ & $\bm{78.96}$ & $\bm{9.74}$ & $\bm{79.8 \pm 4.5}$ \\
\addlinespace[3pt]
\multirow{4}{*}{\textbf{Median}} & Gaussian Grouping & $19.87$ & $0.459$ & $0.319$ & $83.89$ & $75.68$ & $22.30$ & \textemdash \\
 & InFusion & $18.65$ & $0.480$ & $0.316$ & $105.98$ & $128.00$ & $44.25$ & \textemdash \\
 & Inpaint360GS & $19.90$ & $\bm{0.499}$ & $\bm{0.303}$ & $61.02$ & $69.06$ & $9.11$ & \textemdash \\
 & Ours & $\bm{20.02}$ & $0.495$ & $0.307$ & $\bm{51.26}$ & $\bm{64.72}$ & $\bm{6.45}$ & \textemdash \\
        \bottomrule
    \end{tabular}
    }
\end{table*}

%% file: biblio.bib
@article{bigun_multidimensional_1991,
  title = {Multidimensional Orientation Estimation with Applications to Texture Analysis and Optical Flow},
  author = {Bigun, J. and Granlund, G.H. and Wiklund, J.},
  year = 1991,
  month = aug,
  journal = {IEEE Transactions on Pattern Analysis and Machine Intelligence},
  volume = {13},
  number = {8},
  pages = {775--790},
  issn = {1939-3539},
  doi = {10.1109/34.85668},
  urldate = {2026-09-24}
}

@inproceedings{binkowski_demystifying_2018,
  title = {Demystifying {{MMD GANs}}},
  booktitle = {International {{Conference}} on {{Learning Representations}}},
  author = {Bi{\'n}kowski, Miko{\l}aj and Sutherland, Danica J. and Arbel, Michael and Gretton, Arthur},
  year = 2018,
  month = feb,
  urldate = {2026-09-26},
  langid = {english}
}

@article{bradley_rank_1952,
  title = {Rank {{Analysis}} of {{Incomplete Block Designs}}: {{I}}. {{The Method}} of {{Paired Comparisons}}},
  shorttitle = {Rank {{Analysis}} of {{Incomplete Block Designs}}},
  author = {Bradley, Ralph Allan and Terry, Milton E.},
  year = 1952,
  journal = {Biometrika},
  volume = {39},
  number = {3/4},
  eprint = {2334029},
  eprinttype = {jstor},
  pages = {324--345},
  publisher = {[Oxford University Press, Biometrika Trust]},
  issn = {0006-3444},
  doi = {10.2307/2334029},
  urldate = {2026-09-22}
}

@inproceedings{cao_masactrl_2023a,
  title = {{{MasaCtrl}}: {{Tuning-Free Mutual Self-Attention Control}} for {{Consistent Image Synthesis}} and {{Editing}}},
  shorttitle = {{{MasaCtrl}}},
  booktitle = {2023 {{IEEE}}/{{CVF International Conference}} on {{Computer Vision}} ({{ICCV}})},
  author = {Cao, Mingdeng and Wang, Xintao and Qi, Zhongang and Shan, Ying and Qie, Xiaohu and Zheng, Yinqiang},
  year = 2023,
  month = oct,
  pages = {22503--22513},
  issn = {2380-7504},
  doi = {10.1109/ICCV51070.2023.02062},
  urldate = {2026-09-26}
}

@article{cazals_estimating_2005,
  title = {Estimating Differential Quantities Using Polynomial Fitting of Osculating Jets},
  author = {Cazals, F. and Pouget, M.},
  year = 2005,
  month = feb,
  journal = {Computer Aided Geometric Design},
  volume = {22},
  number = {2},
  pages = {121--146},
  issn = {0167-8396},
  doi = {10.1016/j.cagd.2004.09.004},
  urldate = {2026-09-17}
}

@inproceedings{chen_assessing_2024a,
  title = {Assessing {{Image Inpainting}} via {{Re-Inpainting Self-Consistency Evaluation}}},
  booktitle = {Proceedings of the 33rd {{ACM International Conference}} on {{Information}} and {{Knowledge Management}}},
  author = {Chen, Tianyi and Zhang, Jianfu and Hong, Yan and Zhang, Yiyi and Zhang, Liqing},
  year = 2024,
  month = oct,
  pages = {291--300},
  publisher = {ACM},
  address = {Boise ID USA},
  doi = {10.1145/3627673.3679693},
  urldate = {2026-09-26},
  isbn = {979-8-4007-0436-9},
  langid = {english}
}

@inproceedings{chen_gaussianeditor_2024,
  title = {{{GaussianEditor}}: {{Swift}} and {{Controllable 3D Editing}} with {{Gaussian Splatting}}},
  shorttitle = {{{GaussianEditor}}},
  booktitle = {2024 {{IEEE}}/{{CVF Conference}} on {{Computer Vision}} and {{Pattern Recognition}} ({{CVPR}})},
  author = {Chen, Yiwen and Chen, Zilong and Zhang, Chi and Wang, Feng and Yang, Xiaofeng and Wang, Yikai and Cai, Zhongang and Yang, Lei and Liu, Huaping and Lin, Guosheng},
  year = 2024,
  month = jun,
  pages = {21476--21485},
  issn = {2575-7075},
  doi = {10.1109/CVPR52733.2024.02029},
  urldate = {2026-09-26}
}

@inproceedings{chen_gsroadpatching_2025,
  title = {{{GS-RoadPatching}}: {{Inpainting Gaussians}} via {{3D Searching}} and {{Placing}} for {{Driving Scenes}}},
  shorttitle = {{{GS-RoadPatching}}},
  booktitle = {Proceedings of the {{SIGGRAPH Asia}} 2025 {{Conference Papers}}},
  author = {Chen, Guo and Liu, Jiarun and Du, Sicong and Wu, Chenming and Li, Deqi and Huang, Shi-Sheng and Zhang, Guofeng and Yang, Sheng},
  year = 2025,
  month = dec,
  eprint = {2509.19937},
  primaryclass = {cs.CV},
  pages = {1--11},
  doi = {10.1145/3757377.3763892},
  urldate = {2026-09-20},
  archiveprefix = {arXiv}
}

@inproceedings{
chung2023diffusion,
title={Diffusion Posterior Sampling for General Noisy Inverse Problems},
author={Hyungjin Chung and Jeongsol Kim and Michael Thompson Mccann and Marc Louis Klasky and Jong Chul Ye},
booktitle={The Eleventh International Conference on Learning Representations },
year={2023},
url={https://openreview.net/forum?id=OnD9zGAGT0k}
}

@misc{chung_inpaintslat_2026,
  title = {{{InpaintSLat}}: {{Inpainting Structured 3D Latents}} via {{Initial Noise Optimization}}},
  shorttitle = {{{InpaintSLat}}},
  author = {Chung, Jaeyoung and Lee, Suyoung and Lee, Kyoung Mu},
  year = 2026,
  month = may,
  number = {arXiv:2605.00664},
  eprint = {2605.00664},
  primaryclass = {cs.CV},
  publisher = {arXiv},
  doi = {10.48550/arXiv.2605.00664},
  urldate = {2026-09-07},
  archiveprefix = {arXiv}
}

@inproceedings{cicek_3d_2016a,
  title = {{{3D U-Net}}: {{Learning Dense Volumetric Segmentation}} from {{Sparse Annotation}}},
  shorttitle = {{{3D U-Net}}},
  booktitle = {Medical {{Image Computing}} and {{Computer-Assisted Intervention}} -- {{MICCAI}} 2016},
  author = {{\c C}i{\c c}ek, {\"O}zg{\"u}n and Abdulkadir, Ahmed and Lienkamp, Soeren S. and Brox, Thomas and Ronneberger, Olaf},
  editor = {Ourselin, Sebastien and Joskowicz, Leo and Sabuncu, Mert R. and Unal, Gozde and Wells, William},
  year = 2016,
  pages = {424--432},
  publisher = {Springer International Publishing},
  address = {Cham},
  doi = {10.1007/978-3-319-46723-8_49},
  isbn = {978-3-319-46723-8},
  langid = {english}
}

@article{davidson_extending_1977,
  title = {On {{Extending}} the {{Bradley-Terry Model}} to {{Incorporate Within-Pair Order Effects}}},
  author = {Davidson, Roger R. and Beaver, Robert J.},
  year = 1977,
  journal = {Biometrics},
  volume = {33},
  number = {4},
  eprint = {2529467},
  eprinttype = {jstor},
  pages = {693--702},
  publisher = {International Biometric Society},
  issn = {0006-341X},
  doi = {10.2307/2529467},
  urldate = {2026-09-22}
}

@inproceedings{ding_scalable_2025,
  title = {Scalable {{Non-Equivariant 3D Molecule Generation}} via {{Rotational Alignment}}},
  booktitle = {Proceedings of the 42nd {{International Conference}} on {{Machine Learning}}},
  author = {Ding, Yuhui and Hofmann, Thomas},
  year = 2025,
  month = oct,
  pages = {13814--13824},
  publisher = {PMLR},
  issn = {2640-3498},
  urldate = {2026-09-24},
  langid = {english}
}

@article{fischler_random_1981,
  title = {Random Sample Consensus: A Paradigm for Model Fitting with Applications to Image Analysis and Automated Cartography},
  shorttitle = {Random Sample Consensus},
  author = {Fischler, Martin A. and Bolles, Robert C.},
  year = 1981,
  month = jun,
  journal = {Communications of the ACM},
  volume = {24},
  number = {6},
  pages = {381--395},
  issn = {0001-0782, 1557-7317},
  doi = {10.1145/358669.358692},
  urldate = {2026-09-25},
  langid = {english}
}

@inproceedings{he_deep_2016,
  title = {Deep {{Residual Learning}} for {{Image Recognition}}},
  booktitle = {2016 {{IEEE Conference}} on {{Computer Vision}} and {{Pattern Recognition}} ({{CVPR}})},
  author = {He, Kaiming and Zhang, Xiangyu and Ren, Shaoqing and Sun, Jian},
  year = 2016,
  month = jun,
  pages = {770--778},
  publisher = {IEEE},
  address = {Las Vegas, NV, USA},
  doi = {10.1109/CVPR.2016.90},
  urldate = {2026-09-26},
  isbn = {978-1-4673-8851-1}
}

@inproceedings{heusel_gans_2017,
  title = {{{GANs}} Trained by a Two Time-Scale Update Rule Converge to a Local Nash Equilibrium},
  booktitle = {Proceedings of the 31st {{International Conference}} on {{Neural Information Processing Systems}}},
  author = {Heusel, Martin and Ramsauer, Hubert and Unterthiner, Thomas and Nessler, Bernhard and Hochreiter, Sepp},
  year = 2017,
  month = dec,
  series = {{{NIPS}}'17},
  pages = {6629--6640},
  publisher = {Curran Associates Inc.},
  address = {Red Hook, NY, USA},
  urldate = {2026-09-26},
  isbn = {978-1-5108-6096-4}
}

@article{hoppe_surface_1992,
  title = {Surface Reconstruction from Unorganized Points},
  author = {Hoppe, Hugues and DeRose, Tony and Duchamp, Tom and McDonald, John and Stuetzle, Werner},
  year = 1992,
  month = jul,
  journal = {ACM SIGGRAPH Computer Graphics},
  volume = {26},
  number = {2},
  pages = {71--78},
  issn = {0097-8930},
  doi = {10.1145/142920.134011},
  urldate = {2026-09-17},
  langid = {english}
}

@inproceedings{hu_lora_2021a,
  title = {{{LoRA}}: {{Low-Rank Adaptation}} of {{Large Language Models}}},
  shorttitle = {{{LoRA}}},
  booktitle = {International {{Conference}} on {{Learning Representations}}},
  author = {Hu, Edward J. and Shen, Yelong and Wallis, Phillip and {Allen-Zhu}, Zeyuan and Li, Yuanzhi and Wang, Shean and Wang, Lu and Chen, Weizhu},
  year = 2021,
  month = oct,
  urldate = {2026-09-26},
  langid = {english}
}

@inproceedings{huang_3d_2025a,
  title = {{{3D Gaussian Inpainting}} with {{Depth-Guided Cross-View Consistency}}},
  booktitle = {2025 {{IEEE}}/{{CVF Conference}} on {{Computer Vision}} and {{Pattern Recognition}} ({{CVPR}})},
  author = {Huang, Sheng-Yu and Chou, Zi-Ting and Wang, Yu-Chiang Frank},
  year = 2025,
  month = jun,
  pages = {26704--26713},
  issn = {2575-7075},
  doi = {10.1109/CVPR52734.2025.02487},
  urldate = {2026-09-26}
}

@misc{ji_symtrellis_2026,
  title = {{{SymTRELLIS}}: {{Symmetry-Enforced Voxel Latents}} for {{3D Generation}}},
  shorttitle = {{{SymTRELLIS}}},
  author = {Ji, Guangda and Chen, Qimin and Li, Qinchan and Zhao, Mingrui and Wang, Kai and Zhang, Hao},
  year = 2026,
  month = sep,
  number = {arXiv:2606.04108},
  eprint = {2606.04108},
  primaryclass = {cs.GR},
  publisher = {arXiv},
  doi = {10.48550/arXiv.2606.04108},
  urldate = {2026-09-24},
  archiveprefix = {arXiv}
}

@article{kerbl_3d_2023a,
  title = {{{3D Gaussian Splatting}} for {{Real-Time Radiance Field Rendering}}},
  author = {Kerbl, Bernhard and Kopanas, Georgios and Leimkuehler, Thomas and Drettakis, George},
  year = 2023,
  month = aug,
  journal = {ACM Transactions on Graphics},
  volume = {42},
  number = {4},
  pages = {1--14},
  issn = {0730-0301, 1557-7368},
  doi = {10.1145/3592433},
  urldate = {2026-09-26},
  langid = {english}
}

@inproceedings{kheradmand_3d_2024,
  title = {{{3D Gaussian}} Splatting as {{Markov Chain Monte Carlo}}},
  booktitle = {Proceedings of the 38th {{International Conference}} on {{Neural Information Processing Systems}}},
  author = {Kheradmand, Shakiba and Rebain, Daniel and Sharma, Gopal and Sun, Weiwei and Tseng, Yang-Che and Isack, Hossam and Kar, Abhishek and Tagliasacchi, Andrea and Yi, Kwang Moo},
  year = 2024,
  month = dec,
  series = {{{NIPS}} '24},
  volume = {37},
  pages = {80965--80986},
  publisher = {Curran Associates Inc.},
  address = {Red Hook, NY, USA},
  urldate = {2026-09-26},
  isbn = {979-8-3313-1438-5}
}

@inproceedings{kingma_adam_2015,
  title = {Adam: {{A Method}} for {{Stochastic Optimization}}},
  shorttitle = {Adam},
  booktitle = {International {{Conference}} on {{Learning Representations}}},
  author = {Kingma, Diederik P. and Ba, Jimmy},
  year = 2015,
  eprint = {1412.6980},
  primaryclass = {cs.LG},
  publisher = {arXiv},
  doi = {10.48550/arXiv.1412.6980},
  urldate = {2026-08-05},
  archiveprefix = {arXiv}
}

@inproceedings{lee_consistent_2024,
  title = {Consistent {{Object Removal}} from {{Masked Neural Radiance Fields}} by {{Estimating Never-Seen Regions}} in {{All-Views}}},
  booktitle = {Pattern {{Recognition}}: 27th {{International Conference}}, {{ICPR}} 2024, {{Kolkata}}, {{India}}, {{December}} 1--5, 2024, {{Proceedings}}, {{Part XXII}}},
  author = {Lee, Yongjoon and Ryu, Jaehak and Yoon, Donggeun and Cho, Donghyeon},
  year = 2024,
  month = dec,
  pages = {416--431},
  publisher = {Springer-Verlag},
  address = {Berlin, Heidelberg},
  doi = {10.1007/978-3-031-78312-8_28},
  urldate = {2026-09-16},
  isbn = {978-3-031-78311-1}
}

@article{lee_gpgs_2026,
  title = {{{GPGS}}: {{Consistent 3D Object Removal}} via {{Geometry-Aware 3D Inpainting}} and {{Projected Image Refinement}} in {{3D Gaussian Splatting}}},
  shorttitle = {{{GPGS}}},
  author = {Lee, Yongjoon and Cho, Donghyeon},
  year = 2026,
  month = mar,
  journal = {Proceedings of the AAAI Conference on Artificial Intelligence},
  volume = {40},
  number = {8},
  pages = {5927--5935},
  issn = {2374-3468},
  doi = {10.1609/aaai.v40i8.37515},
  urldate = {2026-04-21},
  copyright = {Copyright (c) 2026 Association for the Advancement of Artificial Intelligence},
  langid = {english}
}

@inproceedings{li_voxhammer_2026,
  title = {{{VoxHammer}}: {{Training-Free Precise}} and {{Coherent 3D Editing}} in {{Native 3D Space}}},
  shorttitle = {{{VoxHammer}}},
  booktitle = {2026 {{International Conference}} on {{3D Vision}} ({{3DV}})},
  author = {Li, Lin and Huang, Zehuan and Feng, Haoran and Zhuang, Gengxiong and Chen, Rui and Guo, Chunchao and Sheng, Lu},
  year = 2026,
  month = mar,
  pages = {1281--1292},
  publisher = {IEEE Computer Society},
  doi = {10.1109/3DV69130.2026.00126},
  urldate = {2026-09-26},
  isbn = {979-8-3315-7312-6},
  langid = {english}
}

@misc{liu_infusion_2024a,
  title = {{{InFusion}}: {{Inpainting 3D Gaussians}} via {{Learning Depth Completion}} from {{Diffusion Prior}}},
  shorttitle = {{{InFusion}}},
  author = {Liu, Zhiheng and Ouyang, Hao and Wang, Qiuyu and Cheng, Ka Leong and Xiao, Jie and Zhu, Kai and Xue, Nan and Liu, Yu and Shen, Yujun and Cao, Yang},
  year = 2024,
  month = apr,
  number = {arXiv:2404.11613},
  eprint = {2404.11613},
  primaryclass = {cs},
  publisher = {arXiv},
  doi = {10.48550/arXiv.2404.11613},
  urldate = {2026-04-16},
  archiveprefix = {arXiv}
}

@inproceedings{loshchilov_decoupled_2018,
  title = {Decoupled {{Weight Decay Regularization}}},
  booktitle = {International {{Conference}} on {{Learning Representations}}},
  author = {Loshchilov, Ilya and Hutter, Frank},
  year = 2018,
  month = sep,
  urldate = {2026-09-26},
  langid = {english}
}

@inproceedings{lugmayr_repaint_2022a,
  title = {{{RePaint}}: {{Inpainting}} Using {{Denoising Diffusion Probabilistic Models}}},
  shorttitle = {{{RePaint}}},
  booktitle = {2022 {{IEEE}}/{{CVF Conference}} on {{Computer Vision}} and {{Pattern Recognition}} ({{CVPR}})},
  author = {Lugmayr, Andreas and Danelljan, Martin and Romero, Andres and Yu, Fisher and Timofte, Radu and Van Gool, Luc},
  year = 2022,
  month = jun,
  pages = {11451--11461},
  issn = {2575-7075},
  doi = {10.1109/CVPR52688.2022.01117},
  urldate = {2026-09-26}
}

@inproceedings{mirzaei_spinnerf_2023a,
  title = {{{SPIn-NeRF}}: {{Multiview Segmentation}} and {{Perceptual Inpainting}} with {{Neural Radiance Fields}}},
  shorttitle = {{{SPIn-NeRF}}},
  booktitle = {2023 {{IEEE}}/{{CVF Conference}} on {{Computer Vision}} and {{Pattern Recognition}} ({{CVPR}})},
  author = {Mirzaei, Ashkan and {Aumentado-Armstrong}, Tristan and Derpanis, Konstantinos G. and Kelly, Jonathan and Brubaker, Marcus A. and Gilitschenski, Igor and Levinshtein, Alex},
  year = 2023,
  month = jun,
  pages = {20669--20679},
  publisher = {IEEE},
  address = {Vancouver, BC, Canada},
  doi = {10.1109/CVPR52729.2023.01980},
  urldate = {2026-09-26},
  copyright = {https://doi.org/10.15223/policy-029},
  isbn = {979-8-3503-0129-8}
}

@article{oquab_dinov2_2023,
  title = {{{DINOv2}}: {{Learning Robust Visual Features}} without {{Supervision}}},
  shorttitle = {{{DINOv2}}},
  author = {Oquab, Maxime and Darcet, Timoth{\'e}e and Moutakanni, Th{\'e}o and Vo, Huy V. and Szafraniec, Marc and Khalidov, Vasil and Fernandez, Pierre and Haziza, Daniel and Massa, Francisco and {El-Nouby}, Alaaeldin and Assran, Mido and Ballas, Nicolas and Galuba, Wojciech and Howes, Russell and Huang, Po-Yao and Li, Shang-Wen and Misra, Ishan and Rabbat, Michael and Sharma, Vasu and Synnaeve, Gabriel and Xu, Hu and Jegou, Herve and Mairal, Julien and Labatut, Patrick and Joulin, Armand and Bojanowski, Piotr},
  year = 2023,
  month = jul,
  journal = {Transactions on Machine Learning Research},
  issn = {2835-8856},
  urldate = {2026-09-26},
  langid = {english}
}

@inproceedings{pauly_efficient_2002,
  title = {Efficient Simplification of Point-Sampled Surfaces},
  booktitle = {{{IEEE Visualization}}, 2002. {{VIS}} 2002.},
  author = {Pauly, M. and Gross, M. and Kobbelt, L.P.},
  year = 2002,
  pages = {163--170},
  publisher = {IEEE},
  address = {Boston, MA, USA},
  doi = {10.1109/VISUAL.2002.1183771},
  urldate = {2026-09-17},
  isbn = {978-0-7803-7498-0},
  langid = {english}
}

@misc{pavasovic_wait_2026,
  title = {{{WaiT}} for the {{Signal}}: {{Simple Frequency-Aware Flow-Matching}}},
  shorttitle = {{{WaiT}} for the {{Signal}}},
  author = {Pavasovic, Krunoslav Lehman and Vallaeys, Th{\'e}ophane and Mallat, St{\'e}phane and Biroli, Giulio and Zettlemoyer, Luke and Karrer, Brian and Verbeek, Jakob},
  year = 2026,
  month = jul,
  number = {arXiv:2607.28760},
  eprint = {2607.28760},
  primaryclass = {cs.CV},
  publisher = {arXiv},
  doi = {10.48550/arXiv.2607.28760},
  urldate = {2026-08-10},
  archiveprefix = {arXiv}
}

@misc{sartor_overpainting_2026,
  title = {Overpainting: {{Localized Context-aware Diffusion Image Editing}}},
  shorttitle = {Overpainting},
  author = {Sartor, Sam and Georgiev, Iliyan and Fischer, Michael and Deschaintre, Valentin and Peers, Pieter},
  year = 2026,
  month = sep,
  number = {arXiv:2609.10811},
  eprint = {2609.10811},
  primaryclass = {cs.CV},
  publisher = {arXiv},
  doi = {10.48550/arXiv.2609.10811},
  urldate = {2026-09-24},
  archiveprefix = {arXiv}
}

@inproceedings{song_solving_2023,
  title = {Solving {{Inverse Problems}} with {{Latent Diffusion Models}} via {{Hard Data Consistency}}},
  booktitle = {The {{Twelfth International Conference}} on {{Learning Representations}}},
  author = {Song, Bowen and Kwon, Soo Min and Zhang, Zecheng and Hu, Xinyu and Qu, Qing and Shen, Liyue},
  year = 2023,
  month = oct,
  urldate = {2026-09-26},
  langid = {english}
}

@inproceedings{suvorov_resolutionrobust_2022,
  title = {Resolution-Robust {{Large Mask Inpainting}} with {{Fourier Convolutions}}},
  booktitle = {2022 {{IEEE}}/{{CVF Winter Conference}} on {{Applications}} of {{Computer Vision}} ({{WACV}})},
  author = {Suvorov, Roman and Logacheva, Elizaveta and Mashikhin, Anton and Remizova, Anastasia and Ashukha, Arsenii and Silvestrov, Aleksei and Kong, Naejin and Goka, Harshith and Park, Kiwoong and Lempitsky, Victor},
  year = 2022,
  month = jan,
  pages = {3172--3182},
  issn = {2642-9381},
  doi = {10.1109/WACV51458.2022.00323},
  urldate = {2026-09-26}
}

@misc{tian_3dgimp_2026,
  title = {{{3D-GIMP}}: {{When 3D Gaussian Inpainting Meets PatchMatch}}},
  shorttitle = {{{3D-GIMP}}},
  author = {Tian, Xuening and Schmalstieg, Dieter and Mori, Shohei},
  year = 2026,
  month = jul,
  number = {arXiv:2607.20789},
  eprint = {2607.20789},
  primaryclass = {cs.CV},
  publisher = {arXiv},
  doi = {10.48550/arXiv.2607.20789},
  urldate = {2026-09-20},
  archiveprefix = {arXiv}
}

@inproceedings{wang_inpaint360gs_2026,
  title = {{{Inpaint360GS}}: {{Efficient Object-Aware 3D Inpainting}} via {{Gaussian Splatting}} for 360{$^\circ$} {{Scenes}}},
  shorttitle = {{{Inpaint360GS}}},
  booktitle = {2026 {{IEEE}}/{{CVF Winter Conference}} on {{Applications}} of {{Computer Vision}} ({{WACV}})},
  author = {Wang, Shaoxiang and Zhang, Shihong and Millerdurai, Christen and Westermann, R{\"u}diger and Stricker, Didier and Pagani, Alain},
  year = 2026,
  pages = {117--127},
  publisher = {IEEE},
  urldate = {2026-09-26}
}

@inproceedings{wang_learning_2024,
  title = {Learning {{3D Geometry}} and~{{Feature Consistent Gaussian Splatting}} for~{{Object Removal}}},
  booktitle = {Computer {{Vision}} -- {{ECCV}} 2024: 18th {{European Conference}}, {{Milan}}, {{Italy}}, {{September}} 29--{{October}} 4, 2024, {{Proceedings}}, {{Part III}}},
  author = {Wang, Yuxin and Wu, Qianyi and Zhang, Guofeng and Xu, Dan},
  year = 2024,
  month = oct,
  pages = {1--17},
  publisher = {Springer-Verlag},
  address = {Berlin, Heidelberg},
  doi = {10.1007/978-3-031-72646-0_1},
  urldate = {2026-09-26},
  isbn = {978-3-031-72645-3}
}

@inproceedings{weder_removing_2023,
  title = {Removing {{Objects From Neural Radiance Fields}}},
  booktitle = {2023 {{IEEE}}/{{CVF Conference}} on {{Computer Vision}} and {{Pattern Recognition}} ({{CVPR}})},
  author = {Weder, Silvan and {Garcia-Hernando}, Guillermo and Monszpart, {\'A}ron and Pollefeys, Marc and Brostow, Gabriel and Firman, Michael and Vicente, Sara},
  year = 2023,
  month = jun,
  pages = {16528--16538},
  issn = {2575-7075},
  doi = {10.1109/CVPR52729.2023.01586},
  urldate = {2026-09-26}
}

@inproceedings{wu_amodal3r_2025a,
  title = {{{Amodal3R}}: {{Amodal 3D Reconstruction}} from {{Occluded 2D Images}}},
  shorttitle = {{{Amodal3R}}},
  booktitle = {2025 {{IEEE}}/{{CVF International Conference}} on {{Computer Vision}} ({{ICCV}})},
  author = {Wu, Tianhao and Zheng, Chuanxia and Guan, Frank and Vedaldi, Andrea and Cham, Tat-Jen},
  year = 2025,
  month = oct,
  pages = {9181--9193},
  issn = {2380-7504},
  doi = {10.1109/ICCV51701.2025.00858},
  urldate = {2026-09-26}
}

@inproceedings{wu_aurafusion360_2025a,
  title = {{{AuraFusion360}}: {{Augmented Unseen Region Alignment}} for {{Reference-based}} 360{$^\circ$} {{Unbounded Scene Inpainting}}},
  shorttitle = {{{AuraFusion360}}},
  booktitle = {2025 {{IEEE}}/{{CVF Conference}} on {{Computer Vision}} and {{Pattern Recognition}} ({{CVPR}})},
  author = {Wu, Chung-Ho and Chen, Yang-Jung and Chen, Ying-Huan and Lee, Jie-Ying and Ke, Bo-Hsu and Mu, Chun-Wei Tuan and Huan, Yi-Chuan and Lin, Chin-Yang and Chen, Min-Hung and Lin, Yen-Yu and Liu, Yu-Lun},
  year = 2025,
  month = jun,
  pages = {16366--16376},
  issn = {2575-7075},
  doi = {10.1109/CVPR52734.2025.01526},
  urldate = {2026-09-26}
}

@inproceedings{xiang_structured_2025a,
  title = {Structured {{3D Latents}} for {{Scalable}} and {{Versatile 3D Generation}}},
  booktitle = {2025 {{IEEE}}/{{CVF Conference}} on {{Computer Vision}} and {{Pattern Recognition}} ({{CVPR}})},
  author = {Xiang, Jianfeng and Lv, Zelong and Xu, Sicheng and Deng, Yu and Wang, Ruicheng and Zhang, Bowen and Chen, Dong and Tong, Xin and Yang, Jiaolong},
  year = 2025,
  month = jun,
  pages = {21469--21480},
  issn = {2575-7075},
  doi = {10.1109/CVPR52734.2025.02000},
  urldate = {2026-09-26}
}

@inproceedings{xie_smartbrush_2023,
  title = {{{SmartBrush}}: {{Text}} and {{Shape Guided Object Inpainting}} with {{Diffusion Model}}},
  shorttitle = {{{SmartBrush}}},
  booktitle = {2023 {{IEEE}}/{{CVF Conference}} on {{Computer Vision}} and {{Pattern Recognition}} ({{CVPR}})},
  author = {Xie, Shaoan and Zhang, Zhifei and Lin, Zhe and Hinz, Tobias and Zhang, Kun},
  year = 2023,
  month = jun,
  pages = {22428--22437},
  publisher = {IEEE},
  address = {Vancouver, BC, Canada},
  doi = {10.1109/CVPR52729.2023.02148},
  urldate = {2026-09-21},
  copyright = {https://doi.org/10.15223/policy-029},
  isbn = {979-8-3503-0129-8},
  langid = {english}
}

@inproceedings{yan_generative_2026c,
  title = {Generative {{3D Gaussians}} with {{Learned Density Control}}},
  booktitle = {Proceedings of the {{Special Interest Group}} on {{Computer Graphics}} and {{Interactive Techniques Conference Conference Papers}}},
  author = {Yan, Runjie and Cao, Yan-Pei and Wang, Peng and Liang, Ding and Guo, Yuan-Chen},
  year = 2026,
  month = jul,
  pages = {1--12},
  publisher = {ACM},
  address = {Los Angeles CA USA},
  doi = {10.1145/3799902.3811130},
  urldate = {2026-09-26},
  isbn = {979-8-4007-2554-8},
  langid = {english}
}

@inproceedings{ye_gaussian_2024a,
  title = {Gaussian {{Grouping}}: {{Segment}} and~{{Edit Anything}} in~{{3D Scenes}}},
  shorttitle = {Gaussian {{Grouping}}},
  booktitle = {Computer {{Vision}} -- {{ECCV}} 2024: 18th {{European Conference}}, {{Milan}}, {{Italy}}, {{September}} 29--{{October}} 4, 2024, {{Proceedings}}, {{Part XXIX}}},
  author = {Ye, Mingqiao and Danelljan, Martin and Yu, Fisher and Ke, Lei},
  year = 2024,
  month = nov,
  pages = {162--179},
  publisher = {Springer-Verlag},
  address = {Berlin, Heidelberg},
  doi = {10.1007/978-3-031-73397-0_10},
  urldate = {2026-09-26},
  isbn = {978-3-031-73396-3}
}

@inproceedings{you_instainpaint_2025a,
  title = {{{InstaInpaint}}: {{Instant 3D-Scene Inpainting}} with {{Masked Large Reconstruction Model}}},
  shorttitle = {{{InstaInpaint}}},
  booktitle = {The {{Thirty-ninth Annual Conference}} on {{Neural Information Processing Systems}}},
  author = {You, Junqi and Lin, Chieh Hubert and Lyu, Weijie and Zhang, Zhengbo and Yang, Ming-Hsuan},
  year = 2025,
  month = oct,
  urldate = {2026-09-26},
  langid = {english}
}

@incollection{zhang_gslrm_2025,
  title = {{{GS-LRM}}: {{Large Reconstruction Model}} for {{3D Gaussian Splatting}}},
  shorttitle = {{{GS-LRM}}},
  booktitle = {Computer {{Vision}} -- {{ECCV}} 2024},
  author = {Zhang, Kai and Bi, Sai and Tan, Hao and Xiangli, Yuanbo and Zhao, Nanxuan and Sunkavalli, Kalyan and Xu, Zexiang},
  editor = {Leonardis, Ale{\v s} and Ricci, Elisa and Roth, Stefan and Russakovsky, Olga and Sattler, Torsten and Varol, G{\"u}l},
  year = 2025,
  volume = {15080},
  pages = {1--19},
  publisher = {Springer Nature Switzerland},
  address = {Cham},
  doi = {10.1007/978-3-031-72670-5_1},
  urldate = {2026-09-26},
  isbn = {978-3-031-72669-9 978-3-031-72670-5},
  langid = {english}
}

@misc{zhao_hunyuan3d_2026,
  title = {{{Hunyuan3D}} 2.0: {{Scaling Diffusion Models}} for {{High Resolution Textured 3D Assets Generation}}},
  shorttitle = {{{Hunyuan3D}} 2.0},
  author = {Zhao, Zibo and Lai, Zeqiang and Lin, Qingxiang and Zhao, Yunfei and Liu, Haolin and Yang, Shuhui and Feng, Yifei and Yang, Mingxin and Zhang, Sheng and Yang, Xianghui and Shi, Huiwen and Liu, Sicong and Wu, Junta and Lian, Yihang and Yang, Fan and Tang, Ruining and He, Zebin and Wang, Xinzhou and Liu, Jian and Zuo, Xuhui and Chen, Zhuo and Lei, Biwen and Weng, Haohan and Xu, Jing and Zhu, Yiling and Liu, Xinhai and Xu, Lixin and Hu, Changrong and Yang, Shaoxiong and Zhang, Song and Liu, Yang and Huang, Tianyu and Wang, Lifu and Zhang, Jihong and Chen, Meng and Dong, Liang and Jia, Yiwen and Cai, Yulin and Yu, Jiaao and Tang, Yixuan and Zhang, Hao and Ye, Zheng and He, Peng and Wu, Runzhou and Zhang, Chao and Tan, Yonghao and Xiao, Jie and Tao, Yangyu and Zhu, Jianchen and Xue, Jinbao and Liu, Kai and Zhao, Chongqing and Wu, Xinming and Hu, Zhichao and Qin, Lei and Peng, Jianbing and Li, Zhan and Chen, Minghui and Zhang, Xipeng and Niu, Lin and Wang, Paige and Wang, Yingkai and Kuang, Haozhao and Fan, Zhongyi and Zheng, Xu and Zhuang, Weihao and He, YingPing and Liu, Tian and Yang, Yong and Wang, Di and Liu, Yuhong and Jiang, Jie and Huang, Jingwei and Guo, Chunchao},
  year = 2026,
  month = may,
  number = {arXiv:2501.12202},
  eprint = {2501.12202},
  primaryclass = {cs.CV},
  publisher = {arXiv},
  doi = {10.48550/arXiv.2501.12202},
  urldate = {2026-09-04},
  archiveprefix = {arXiv}
}

@article{zhou_highfidelity_2025a,
  title = {High-Fidelity {{3D Gaussian}} Inpainting: {{Preserving}} Multi-View Consistency and Photorealistic Details},
  shorttitle = {High-Fidelity {{3D Gaussian}} Inpainting},
  author = {Zhou, Jun and Li, Dinghao and Li, Nannan and Wang, Mingjie},
  year = 2025,
  month = oct,
  journal = {Computers \& Graphics},
  volume = {131},
  pages = {104362},
  issn = {00978493},
  doi = {10.1016/j.cag.2025.104362},
  urldate = {2026-09-26},
  langid = {english}
}

@article{
    xiang2025trellis2,
    title={Native and Compact Structured Latents for 3D Generation},
    author={Xiang, Jianfeng and Chen, Xiaoxue and Xu, Sicheng and Wang, Ruicheng and Lv, Zelong and Deng, Yu and Zhu, Hongyuan and Dong, Yue and Zhao, Hao and Yuan, Nicholas Jing and Yang, Jiaolong},
    journal={Tech report},
    year={2025}
}
